\documentclass[lettersize,journal]{IEEEtran}

\usepackage{color}
\usepackage{xcolor}
\usepackage{multirow}
\usepackage{indentfirst}
\usepackage{csquotes}
\usepackage{adjustbox}
\usepackage{enumitem}
\usepackage{booktabs}
\usepackage{subcaption}

\usepackage{amsmath,amsfonts}
\usepackage{algorithmic}
\usepackage{array}
\usepackage[caption=false,font=normalsize,labelfont=sf,textfont=sf]{subfig}
\usepackage{textcomp}
\usepackage{stfloats}
\usepackage{url}
\usepackage{verbatim}
\usepackage{graphicx}
\def\BibTeX{{\rm B\kern-.05em{\sc i\kern-.025em b}\kern-.08em
    T\kern-.1667em\lower.7ex\hbox{E}\kern-.125emX}}
\usepackage{balance}
\begin{document}
\title{MV-CLIP: Multi-View CLIP for Zero-shot 3D Shape Recognition}
\author{Dan Song, Xinwei Fu, Ning Liu, Weizhi Nie, Wenhui Li, Lanjun Wang,\\You Yang~\IEEEmembership{Senior Member,~IEEE},
Anan Liu*~\IEEEmembership{Senior Member,~IEEE}  

\IEEEcompsocitemizethanks{
\IEEEcompsocthanksitem This work was supported by the National Natural Science Foundation of China under Grant U21B2024, Grant U22A2068, and Grant 62272337, Grant 62202327.
\IEEEcompsocthanksitem  Dan Song, Xinwei Fu, Ning Liu, Weizhi Nie, Wenhui Li and An-An Liu are with the School of Electrical and Information Engineering, Tianjin University, Tianjin 300072, China. 
\IEEEcompsocthanksitem Lanjun Wang is with  the School of New Media and Communication, Tianjin University, Tianjin 300072, China. 
\IEEEcompsocthanksitem You Yang is with the School of Electronic Information and Communications, Huazhong University of Science and Technology, Wuhan 430074, China.
\IEEEcompsocthanksitem	Corresponding author: An-An Liu, anan0422@gmail.com.}
}

\maketitle

\begin{abstract}
Large-scale pre-trained models have demonstrated impressive performance in vision and language tasks within open-world scenarios.
Due to the lack of comparable pre-trained models for 3D shapes, recent methods utilize language-image pre-training to realize zero-shot 3D shape recognition.
However, due to the modality gap, pretrained language-image models are not confident enough in the generalization to 3D shape recognition.
Consequently, this paper aims to improve the confidence with view selection and hierarchical prompts.
Leveraging the CLIP model as an example, we employ view selection on the vision side by identifying views with high prediction confidence from multiple rendered views of a 3D shape.
On the textual side, the strategy of hierarchical prompts is proposed for the first time. The first layer prompts several classification candidates with traditional class-level descriptions, while the second layer refines the prediction based on function-level descriptions or further distinctions between the candidates.
Remarkably,  without the need for additional training, our proposed method achieves impressive zero-shot 3D classification accuracies of 84.44\%, 91.51\%, and 66.17\% on ModelNet40, ModelNet10, and ShapeNet Core55, respectively. Furthermore, we will make the code publicly available to facilitate reproducibility and further research in this area.
\end{abstract}


\section{Introduction}
\IEEEPARstart{W}{ith} the extensive applications of 3D models in computer-aided design (CAD), autonomous driving and virtual reality/augmented reality (VR/AR), together with the rapid advancements in 3D scanning and reconstruction technologies, the number of 3D shapes has experienced an explosive increase. How to effectively identify and manage these unlabeled  3D data has become a challenging problem \cite{tcsvt_li2023focus,tcsvt_su2019joint,tcsvt_sun2021joint}. Zero-shot 3D shape recognition aims to classify unseen 3D shapes without explicit training, which has become a hot topic in computer vision with significant benefits such as identifying novel objects and alleviating labor-intensive annotations.

Traditional zero-shot methods \cite{3D_zero1,3D_zero2,3D_zero3} rely on a limited distribution of ``seen'' 3D shape data, resulting in insufficient generalization to new ``unseen'' categories. Additionally, the hand-crafted semantic attributes designed for ``seen'' data cannot cover the characteristics of ``unseen'' data and mapping high-level shape features to these attributes is difficult.
Pre-trained large-scale models have demonstrated strong generalization capabilities, making them highly favored for zero-shot tasks. Due to the absence of comparable pre-trained models specifically designed for 3D shapes, recent methods utilize vision-language models (e.g., CLIP \cite{13_clip}) to realize zero-shot 3D shape recognition, which can be classified into training-based and non-training-based methods. Training-based methods \cite{14_xue2023ulip,16_hegde2023cg3d} create multi-modal datasets for 3D shapes and perform multi-modal contrasts. Integrating 3D data into the pre-training stage greatly enhances the capabilities of zero-shot 3D recognition. However, such approach comes with challenges such as the need for large-scale 3D data, extensive pre-processing requirements, and high computational training costs. Non-training-based methods \cite{01_zhang2022pointclip,03_zhu2022pointclip,23_shen2023diffclip} render 3D shapes into images which are later encoded by the visual encoder of CLIP and compared with the textual encoding of category labels. Our approach falls into the non-training paradigm and raises concerns about the reliability of CLIP when applied to 3D shape recognition, considering both the visual and textual prompt aspects.

\begin{figure}[t!]
    \centering
    \includegraphics[width=1\linewidth]{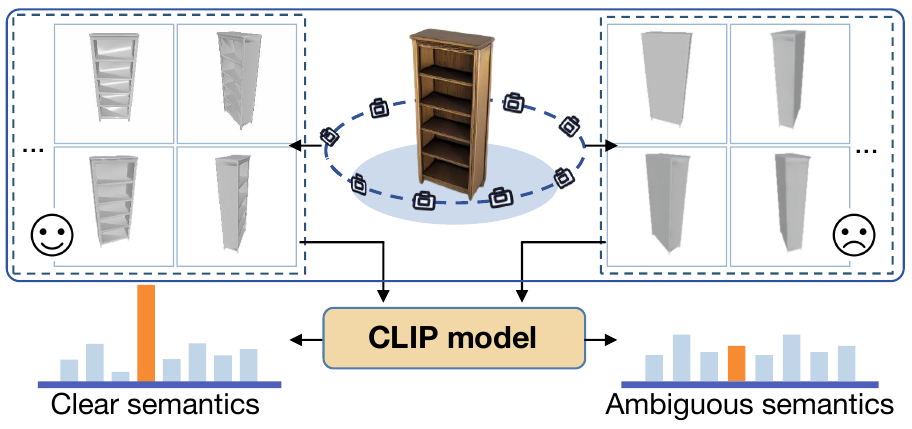}
    \caption{Improve CLIP's confidence at zero-shot 3D shape recognition in the visual aspect: select images with clear semantics.}
    \label{fig:01_vs_importance}
\end{figure}

\textbf{Visual aspect.} 
PointCLIP \cite{01_zhang2022pointclip} projected 3D point clouds to sparsely distributed points in a depth map, which for the first time leverage CLIP for zero-shot 3D shape recognition. To improve the image quality according to CLIP's preference,  PointCLIP V2 \cite{03_zhu2022pointclip} transformed point clouds into voxels to generate much smoother projection values and DiffCLIP \cite{23_shen2023diffclip} enhanced the style of depth map closely to natural photos with diffusion model. CLIP2Point \cite{02_huang2023clip2point} enhanced the projection between points and pixels and refined the issue of excessive blank area in rendered depth images.
However, as shown in Figure~\ref{fig:01_vs_importance}, the view images with ambiguous semantics confuse CLIP and will hinder the performance of shape recognition.

\begin{figure}[t!]
    \centering
    \includegraphics[width=1\linewidth]{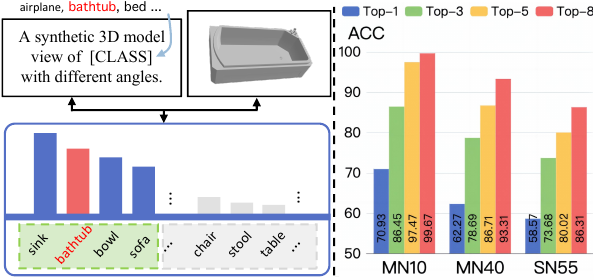}
    \caption{Improve CLIP's confidence at zero-shot 3D shape recognition in the textual prompt: refine the prediction with hierarchical prompts. Left: An example of bathtub that is mis-classified into sink. Right: Statistical zero-shot top-k accuracy on three popular datasets, which is obtained by the CLIP model under the settings of 12 pre-defined online-rendered views and hand-crafted prompts of first layer.}
    \label{fig:01_hp_importance}
\end{figure}

\textbf{Prompt aspect.}
PointCLIP \cite{01_zhang2022pointclip} adopted a hand-crafted template ``{point cloud depth map of a [class]}" as prompt. Similarly, DiffCLIP \cite{23_shen2023diffclip} used ``{a 3D rendered image of a [class]}". To design a more detailed prompt in 3D perspective, PointCLIP V2 fed 3D command into GPT-3 like ``{Give a caption of a table depth map}" and obtained 3D-specific prompt such as ``{A height map of a table with a top and several legs}". By matching the visual feature and textual feature encoded by pre-trained model, the class that owns the highest similarity score becomes the predicted result.
However, based on our observations as shown in Figure~\ref{fig:01_hp_importance}, sometimes the visual encoding does not perfectly match the ground-truth textual encoding, but usually ranks at the forefront. Directly adopting the top-1 result as prediction will limit further optimization.  

To improve the confidence of pre-trained vision-language models towards the task of zero-shot 3D shape recognition, the proposed MV-CLIP (Multi-View CLIP) is equipped with view selection and hierarchical prompts. On the visual side, we select the view images with clear semantics based on the prediction entropy and fuse the classification results of the selected views as the final prediction. On the textual side, we propose a novel prompt mechanism named hierarchical prompts. For the first layer, we design a hand-crafted prompt as ``{A synthetic 3D model view of [class] with different angles}". By matching the encoding of selected views and hand-crafted prompts via MV-CLIP, we acquire top-k results as candidates for further consideration. The second layer for refinement is designed based on current large language model (GPT-3.5 \cite{04_brown2020language}). Specifically towards the candidate classes, we feed sentences like ``{Describe the visual characteristics and functional features of [candidate classes]}'' or ``{What is the difference between [candidate classes] in visual characteristics and functional features}". The concise answer generated by GPT is then used as the textual prompts for these candidates. The hand-crafted part is a uniform template which is not affected by classes, and we do not require any pre-training.  The candidate classes can be easily and dynamically altered, offering a certain level of openness.

View selection improves CLIP's confidence by filtering confusing views while hierarchical prompts give MV-CLIP a second chance with more specific descriptions.
Consequently, without any training the proposed method achieves impressive zero-shot 3D classification accuracies of 84.44\%, 91.51\%, and 66.17\% on ModelNet40, ModelNet10, and ShapeNet Core55, respectively. With the performance improvement of vision-language models, the proposed method can be further extended to the latest models to enhance zero-shot shape recognition performance.

In summary, the contributions are as follows:
\begin{itemize}[leftmargin=0.3cm]
\item We propose MV-CLIP for directly extracting multi-view features. We utilize the pre-trained image encoder as the backbone of the multi-view network, achieving state-of-the-art performance in zero-shot 3D shape recognition.
\item We introduce a view selection module to evaluate the quality of each view based on the principle of entropy minimization. This allows us to identify views that have a positive impact on MV-CLIP.
\item We propose a novel hierarchical prompts strategy to improve the matching between MV-CLIP's view representations and textual prompts. With the candidates voted by the first-layer classification, LLMs-powered prompts towards candidates contribute more accurate second-layer matching.
\end{itemize}

\section{Related Work}

\subsection{Zero-shot Learning in 3D Shape Recognition}
In recent years, Vision-Language Models (VLMs) such as CLIP \cite{13_clip}, which employs large-scale image-text contrastive pre-training, have achieved remarkable success in the realm of 2D visual task. Many works \cite{29_wang2023beyond,30_sanghi2023sketch,31_zhang2023clip,17_zeng2023clip2,01_zhang2022pointclip,02_huang2023clip2point,14_xue2023ulip,16_hegde2023cg3d,18_liu2023openshape} have explored to apply VLMs in 3D learning, thereby attaining the capability of zero-shot recognition.

Some methods directly apply VLMs to the 3D domain.
PointCLIP \cite{01_zhang2022pointclip} stands as the pioneering effort in applying VLMs to 3D recognition. It directly projects point clouds into multi-view depth maps and utilizes a frozen CLIP model for zero-shot classification. 
With this advancement, PointCLIP V2 \cite{03_zhu2022pointclip} introduces more realistic shape projection and utilizes LLMs-assisted 3D prompts to effectively mitigate the 2D-3D domain gap. Meanwhile, CLIP2Point \cite{02_huang2023clip2point} fixes CLIP  and additionally trains a depth encoder using 3D dataset with initialization of CLIP's  visual encoder.

In addition, other methods explore extending the multi-modal learning between images and language to 3D modalities \cite{14_xue2023ulip,15_xue2023ulip,16_hegde2023cg3d,17_zeng2023clip2,18_liu2023openshape}, achieving impressive zero-shot 3D model recognition performance. ULIP \cite{14_xue2023ulip} learns a unified representation between language, images, and point clouds, and it significantly enhances the recognition capability of 3D backbone. 
Furthermore, ULIP-2 \cite{15_xue2023ulip} specifically focuses on the scalability and comprehensiveness of the language modality. The more recent work of Openshape \cite{18_liu2023openshape} also adopts a multi-modality contrastive learning framework, which improves the ability of open-world 3D shape understanding by improving aspects such as data scalability, text quality, 3D backbone scaling, and data resampling.

In this paper, we focus on zero-shot 3D recognition without the necessity for 3D pre-training, resulting in outstanding zero-shot classification performance, comparable to the state-of-the-art results achieved by multi-modal contrast methods that need fine-tuning.

\subsection{Multi-View in 3D Shape Recognition}
For 3D shape recognition, multi-view representation stands as one of the most classical types, which employs 2D views from multiple perspectives to represent 3D shapes \cite{tcsvtr_huang2021learning, 08_wei2020view,01_zhang2022pointclip}. 
The work proposed by Bradski et al. \cite{05_bradski1994recognition} was the trailblazer in employing multiple views to depict 3D shapes. Subsequently, with the advancement of deep learning, MVCNN \cite{06_su2015multi} uses 2D CNN to extract features from a set of predefined views and aggregates them into a descriptor that could effectively represent 3D shapes.
View-GCN \cite{08_wei2020view} utilizes dynamic graph convolutional networks for hierarchical learning, leveraging inter-view relationship information to aggregate multi-view features. MVTN \cite{09_hamdi2021mvtn} combines differentiable rendering techniques for predicting the optimal viewing angles in multi-view setups, which enhances the robustness and recognition performance of multi-view networks. Additionally, some recent studies \cite{01_zhang2022pointclip,10_goyal2021revisiting,11_zhang2022point,12_zhang2023learning} process 3D point clouds in the form of multiple depth images, which further emphasizes the importance of multi-view in 3D shape recognition.

However, it is essential to ensure the effectiveness of each view in the multiple views for subsequent tasks.
In this study, we exploit the extensive semantic insights provided by VLMs \cite{13_clip} to assess the efficacy of multiple views. We select a subset of views for each 3D shape, prioritizing those that offer semantic clarity and excluding those that introduce ambiguous information.

\subsection{Prompt Learning in Vision}
The concept of prompts is initially introduced in the field of Natural Language Processing (NLP) \cite{37_liu2023pre,38_jiang2020can,39_wallace2019universal}, and has found increasingly flexible and widespread application in pre-trained language models such as BERT \cite{36_devlin2018bert} and the GPT \cite{04_brown2020language} series. 
Inspired by the success of prompts in NLP, the practice of prompt engineering has also been adopted in 2D vision \cite{13_clip,41_zhou2022conditional,42_zhou2022learning,43_khattak2023maple,44_jia2022visual,45_bahng2022visual,46_guo2023viewrefer}. Some of them employ a few learnable prompts either in the encoder input \cite{41_zhou2022conditional,42_zhou2022learning} or within the transformer layers \cite{43_khattak2023maple} to adapt the model for enhanced alignment between text and images. Others introduce visual prompting to apply in the pixel space \cite{44_jia2022visual,45_bahng2022visual} or embeddings of input images \cite{46_guo2023viewrefer} without extensive retraining or fine-tuning.

Additionally, in vision approaches related to CLIP \cite{13_clip}, notable advancements have been achieved by integrating LLMs to refine the prompts. CuPL \cite{48_pratt2023does} and CaFo \cite{47_zhang2023prompt} utilize GPT-3 \cite{04_brown2020language} to improve the downstream capabilities of CLIP in handling a variety of 2D datasets.
Meanwhile, CHiLS \cite{49_novack2023chils} employs GPT-3 \cite{04_brown2020language} to generate subclass labels that form mutually mapping hierarchical label sets, which are utilized to produce the final prediction.
PointCLIP V2 \cite{03_zhu2022pointclip} prompts GPT-3 \cite{04_brown2020language} to enhance performance in open-world 3D tasks using 3D-oriented commands. Concurrently, ULIP-2 \cite{15_xue2023ulip} and OpenShape \cite{18_liu2023openshape} leverage LLMs to improve the quality of 3D prompts, aiming to a more effective alignment across various modalities. 
Although LLMs facilitate the design of prompts, the single-step top-1 prediction limits further improvement.

In this paper, we propose a novel strategy of hierarchical prompts. At the first layer, candidate categories are voted via matching hand-crafted prompts. At the second layer, we further utilize GPT-3.5 \cite{04_brown2020language} to generate 3D-specific prompts for these candidates in aspects of function and difference.
The design of hierarchical prompts enables more accurate category prediction without the necessity for training.

\section{Methodology}
\begin{figure*}[t!]
	\centering
 	\includegraphics[width=0.87\textwidth]{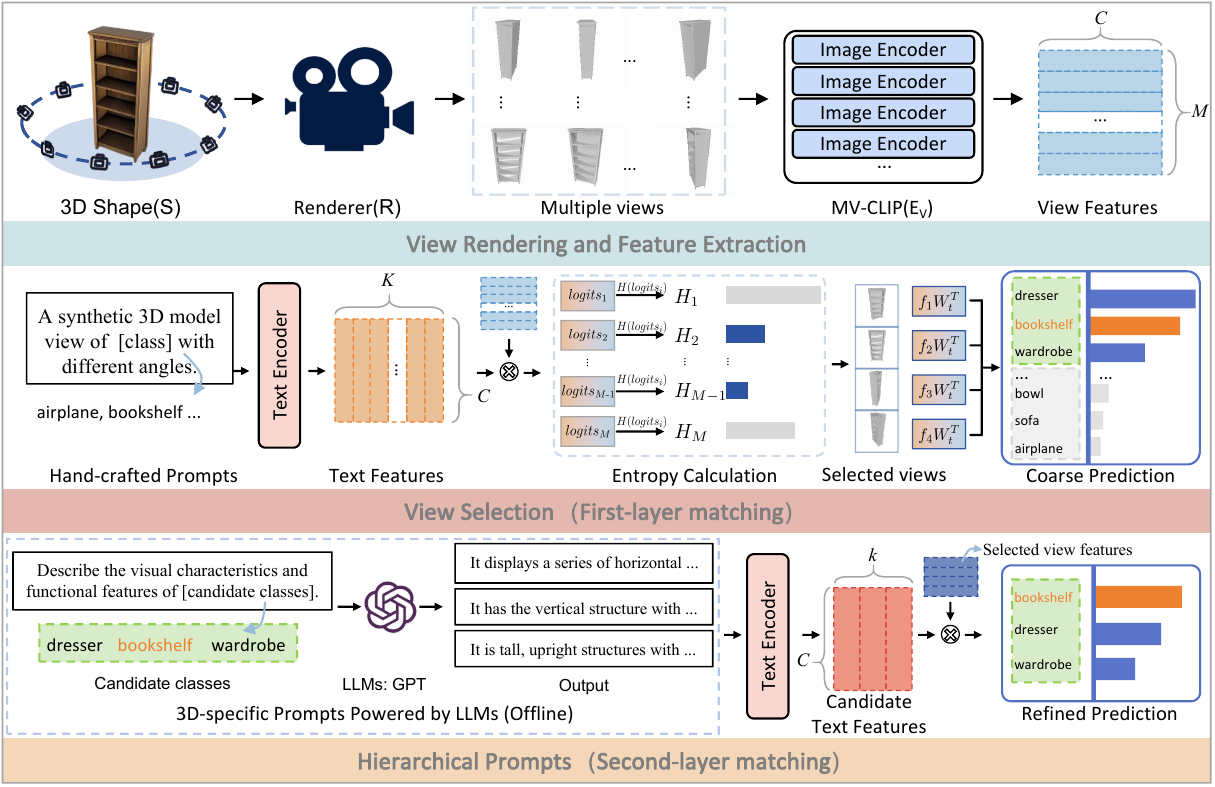}
	\caption{Overview of the proposed MV-CLIP for zero-shot 3D shape recognition. Firstly, multiple view images are obtained via a render $\mathbf{R}$ and the corresponding visual features are extracted via the visual encoder of CLIP \cite{13_clip}. Secondly, visual features are matched with textual features encoded by CLIP with hand-crafted prompts, and we select representative views according the prediction confidence. By aggregating the representative predictions, several candidates with the top classification probability are kept for the second matching. Finally, by matching the prompts powered by LLMs  for these candidates, the prediction result is refined.}
	\label{fig:overview}      
\end{figure*}

The overview of Multi-View CLIP (MV-CLIP) for zero-shot 3D shape recognition is illustrated in Figure~\ref{fig:overview}. 
In Sec. \ref{sec:3_1}, we give a brief introduction to multi-view rendering and visual feature extraction. Sec. \ref{sec:3_2} explains how semantic information indicated by CLIP \cite{13_clip} is utilized to filter ambiguous views. Furthermore, in Sec. \ref{sec:3_3}, we design hierarchical prompts to firstly propose candidates and then refine the prediction.

\subsection{View Rendering and Feature Extraction}
\label{sec:3_1}
In order to fully validate and implement the proposed view selection module, we employ a multi-view online renderer \(\mathbf{R}\) \cite{19_hamdi2022mvtn}.
In our approach, we initially render a total of $M$ views $X=\left \{ x_{i} \right \} _{i=1}^M$ for a 3D shape based on fixed view configurations. We adopt three different view configurations, among which the circular \cite{06_su2015multi} aligns view-points on a circle around the object, the spherical \cite{07_kanezaki2018rotationnet,08_wei2020view} aligns equally spaced view-points on a sphere surrounding the object, and the random selects randomly view-points around the object.
In addition, we replace 2D CNNs in the traditional multi-view convolutional neural network \cite{06_su2015multi} with pre-trained visual encoder, which is referred to as MV-CLIP for zero-shot 3D recognition network. The feature extraction for multiple views is formulated as:
\begin{equation}
    \left \{ f \right \} _{i=1}^{M}  = \mathbf{E}_{V}\left ( \mathbf{R}(\mathbf{S}) \right ) 
\label{eq:feature_ex}
\end{equation}
where $\mathbf{E}_{V}$ denotes the visual encoder of MV-CLIP, \(\mathbf{R}\) represents the online renderer, and \(\mathbf{S}\) stands for an arbitrary 3D shape.

\subsection{View Selection}
\label{sec:3_2}

The objective of this module is to select views that capture clear semantic features of the 3D shape from multiple pre-defined views.
As entropy reflects the prediction uncertainty, we utilize it to evaluate the prediction confidence of pre-trained models.
The views with higher prediction confidence are selected as representative views. In the following contents, we will first elaborate the prediction process via CLIP \cite{13_clip} and then compute entropy for view selection.

Specifically, we evaluate the semantic representation of each view by matching the straightforward and hand-crafted prompts.
Based on Eq.~\ref{eq:feature_ex}, multiple views are passed through MV-CLIP to obtain corresponding visual features, denoted as $\left \{ f \right \} _{i=1}^{M} \in \mathbb{R} ^{M\times C} $. 
For textual prompts, we design a pre-defined template: ``{A synthetic 3D model view of [class] with different angles .}'' to generate hand-crafted prompts containing $K$ categories and encode their textual features as $W_{t}\in \mathbb{R} ^{K\times C} $.
Subsequently, the prediction of each view is calculated separately,
\begin{equation}
    logits_{i} = f_{i}W_{t}^{T},for\; i=1,\dots ,M
\label{eq:classi_view}
\end{equation}
where each bit of $logits_{i}$ represents the similarity score between the $i^{th}$ view and each category.

Then the entropy of $logits_{i}$ is computed as:

\begin{equation}
    H(logits_{i} )= - \sum_{j=1}^{K}P(logits_{i,j} ) \log_{2}{P(logits_{i,j})} 
\label{eq:entropy}
\end{equation}
where $H(\cdot)$ denotes the entropy of $logits_{i}$, which is the information summation over all possible categories. Besides, $P(logits_{i,j})$ represents the probability of the occurrence of the $j^{th}$ category, and the term $\log_{2}{P(logits_{i,j})}$ is used to quantify the information content of the probability.
Lower entropy signifies higher information quality compared with other views of the same 3D shape.

We rank all views by the entropy of prediction and select a subset of views with clearer semantics, i.e., smaller entropy, to represent the 3D shape. The selected views are denoted as:
\begin{equation}
X_{selec}  = \left \{ x_{i} | Rank(H(logits_{i} ) ) <= M_{selec} \right \}  \\
\label{eq:selected}
\end{equation}
where the $Rank(\cdot)$ is used to sort the information quality in ascending order, and $M_{selec}$ represents the number of selected views.

\begin{table*}[thb]
    \centering
        \caption{ Zero-shot 3D shape classification performance. We compare the experimental results of existing zero-shot 3D learning methods using their best-performing settings on ModelNet10 and ModelNet40.}
        \begin{tabular}{ccccc}
            \toprule
            \multirow{2}{*}{Method}                                              & \multirow{2}{*}{CLIP version}        & \multirow{2}{*}{Pre-training source}                       & \multicolumn{2}{c}{Zero-shot performance} \\ \cmidrule{4-5} 
                                                                                 &                                      &                                                            & ModelNet40          & ModelNet10          \\
            \midrule
            CG3D\cite{16_hegde2023cg3d}+PointTransformer\cite{20_zhao2021point}  &\multirow{4}{*}{SLIP\cite{28_mu2022slip}}                 & ShapeNet\cite{24_chang2015shapenet}                        & 50.6                & -                   \\
            ULIP\cite{14_xue2023ulip}+PointBERT\cite{21_yu2022point}             &                                      & ShapeNet\cite{24_chang2015shapenet}                        & 60.4                & -                   \\
            ULIP2\cite{15_xue2023ulip}+Point-BERT\cite{21_yu2022point}           &                                      & ShapeNet\cite{24_chang2015shapenet}                        & 66.4                & -                   \\ 
            ULIP2\cite{15_xue2023ulip}+Point-BERT\cite{21_yu2022point}           &                                      & Objaverse\cite{25_deitke2023objaverse}                     & 74                  & -                   \\ \cline{2-2} 
            OpenShape\cite{18_liu2023openshape}+PointBERT\cite{21_yu2022point}   & \multirow{2}{*}{OpenCLIP\cite{27_ilharco2openclip}}            & ShapeNet\cite{24_chang2015shapenet}                        & 72.9                & -                   \\
            OpenShape\cite{18_liu2023openshape}+PointBERT\cite{21_yu2022point}   &                                      & Ensembled(no LVIS)\cite{18_liu2023openshape}               & \textbf{85.3}       & -                   \\
            \midrule
            CLIP2Point\cite{02_huang2023clip2point}                              & \multirow{2}{*}{CLIP\cite{13_clip}}                &\multirow{2}{*}{ShapeNet\cite{24_chang2015shapenet}}        & 49.38               & 66.63               \\
            Recon\cite{22_qi2023contrast}                                        &                                      &                                                            & 61.7                & 75.6                \\
            \midrule
            PointCLIP\cite{01_zhang2022pointclip}                                & \multirow{3}{*}{CLIP\cite{13_clip}}                & $\times $                                                          & 20.18               & 30.23               \\
            PointCLIP v2\cite{03_zhu2022pointclip}                               &                                      & $\times $                                                          & 64.22               & 73.13               \\
            DiffCLIP\cite{23_shen2023diffclip}                                  &                                      & $\times $                                                          & 49.7                & 80.6                \\
            \hline
            \multirow{2}{*}{Ours}                                               &CLIP\cite{28_mu2022slip}                                  & $\times $                                                          & 65.92               & 77.53               \\ 
                                                                                &OpenCLIP\cite{27_ilharco2openclip}                              & $\times $                                                          & 84.44               & \textbf{91.51}      \\
            \bottomrule
    \end{tabular}

    \label{tab4:zero_shotacc}
\end{table*}

\subsection{Hierarchical Prompts}
\label{sec:3_3}

\subsubsection{3D-Specific Prompts Powered by LLMs}
As previously shown in Figure~\ref{fig:01_hp_importance}, due to modality gap, the prediction via CLIP is not confident enough. Therefore, we propose the hierarchical prompts where for the first layer hand-crafted prompts with clear category indication are used to vote several class candidates and for the second layer we utilize the prompts generated by powerful LLMs for these candidate classes. The prediction result gets refined by twice matching the view feature with hierarchical prompts.

With the consideration that the pre-trained models is trained using a collection of image-text pairs obtained from the Internet, besides focusing on the difference between candidates, we enhance the richness of prompts for augmenting the functional attributes. 
Specifically, we employ a pre-defined question template and utilize the GPT-3.5 \cite{04_brown2020language} to generate 3D textual description: ``{Describe the visual characteristics and functional features of [candidate classes]'s rendering view.}''

Take the candidate classes as [dresser, bookshelf, wardrobe] for example, GPT-3.5 produces: 
\begin{itemize}[leftmargin=0.3cm]
    \item {It displays a series of horizontal shelves designed to hold books to provide storage for reading materials.}
    \item {It has the vertical structure with multiple drawers used for storing clothes or personal items.} 
    \item {It is tall, upright structures with multiple compart- ments or shelves for storing clothes and accessories.}
\end{itemize}

Subsequently, the specific prompts are encoded by the pre-trained textual encoder $\mathbf{E}_{T}$ into text features, which are then utilized for second-layer classification.
Formally, the textual features for candidate classes are represented as:
\begin{equation}
 \left \{t \right \} _{i=1}^{k} = \mathbf{E}_{T}(\mathbf{LLM}(Q_{i}))
  \label{eq:second_text}
\end{equation}
where $Q_{i}$ is the question template with the $i^{th}$ candidate class and $k$ represents the number of candidate classes.

\subsubsection{Hierarchical Prompts Matching}

The potential of the matching capability of 3D shape using hand-crafted prompts has been overlooked in zero-shot 3D learning with CLIP. 
Under the hand-crafted prompts, the first-layer classification score via MV-CLIP is computed by aggregating the scores of the selected views. Formally, denote the prediction of a 3D shape in the first layer as:
\begin{equation}
logits^{\uppercase\expandafter{\romannumeral1} } =  {\textstyle \sum_{i=1}^{M_{selec} }logits_{i} } 
  \label{eq:first_layer}
\end{equation}
where $logits_{i}$ is computed as Eq.~\ref{eq:classi_view}, and $M_{selec}$ denotes the number of selected views.

If the prediction output by the first layer is not confident enough, e.g., the maximum probability within $logits^{\uppercase\expandafter{\romannumeral1} }$ is lower than a threshold $\delta$, we choose the $top\text{-}k$ labels as candidate classes for the second matching.
Subsequently, at the second-layer, we employ the specific prompts generated by LLMs for re-matching to attain optimal classification outcomes.

Suppose the textual features in the second layer towards candidate classes are $W_{t_{k}} = \left \{t_{1}; t_{2}; t_{3}; \cdots; t_{k}\right \}\in \mathbb{R} ^{k\times C} $ where $t$ is encoded as Eq.~\ref{eq:second_text}, the final prediction is computed as:   
\begin{equation}
logits^{\uppercase\expandafter{\romannumeral2} } =  {\textstyle \sum_{i=1}^{M_{selec} }f_{i}W_{t_{k}}^{T} } 
\label{eq:important}
\end{equation}

\begin{equation}
y_{pre} = Argmax(logits^{\uppercase\expandafter{\romannumeral2} } )
  \label{eq:also-important}
\end{equation}

Rather than directly classifying within the initial categories, we combine hand-crafted and LLMs generated prompts into a novel strategy of hierarchical prompts. For less certain samples, MV-CLIP will further focus on category details within a limited set of candidate labels, thereby achieving more accurate zero-shot recognition results.

\begin{table*}[ht]
    \centering
    \caption{Performance comparison with different components. We conduct ablation study on ModelNet10, ModelNet40 and ShapeNet Core55 to explore the impact of individual designed modules on the experimental results, respectively. }
    \begin{tabular}{ccccccc}
    \toprule
    \multirow{2}{*}{\begin{tabular}[c]{@{}c@{}}CLIP\\ (ViT$\backslash$B-16)\end{tabular}} & \multirow{2}{*}{\begin{tabular}[c]{@{}c@{}}OpenCLIP\\ (ViT$\backslash$B-16)\end{tabular}} & \multirow{2}{*}{\begin{tabular}[c]{@{}c@{}}View\\ selection\end{tabular} } & \multirow{2}{*}{\begin{tabular}[c]{@{}c@{}}Hierarchical\\ prompts\end{tabular} } & \multicolumn{3}{c}{Zero-shot 3D classification}   \\ \cmidrule{5-7} 
             &              &           &                & ModelNet40     & ModelNet10     & ShapeNet Core55 \\
    \midrule
    $\surd$         & $\times $           & $\times $        & $\times $             & 61.93          & 70.70           & 57.90            \\
    $\surd$         & $\times $           & $\times $        & $\surd$               & 63.85 \textcolor[HTML]{138906}{\small ($\uparrow 1.92$)}           & 71.37 \textcolor[HTML]{138906}{\small ($\uparrow 0.67$)}           & 58.70 \textcolor[HTML]{138906}{\small ($\uparrow 0.8$)}             \\
    $\surd$         & $\times $           & $\surd$          & $\times $             & 64.18 \textcolor[HTML]{138906}{\small ($\uparrow 2.25$)}           & 76.34 \textcolor[HTML]{138906}{\small ($\uparrow 5.64$)}           & 60.80 \textcolor[HTML]{138906}{\small ($\uparrow 2.9$)}             \\
    $\surd$         & $\times $           & $\surd$          & $\surd$               & 65.92  \textcolor[HTML]{138906}{\small ($\uparrow 3.99$)}          & 77.53 \textcolor[HTML]{138906}{\small ($\uparrow 6.83$)}           & 61.70 \textcolor[HTML]{138906}{\small ($\uparrow 3.8$)}             \\

    $\times $       & $\surd$             & $\times $        & $\times $             & 78.03          & 86.45          & 60.62           \\
    
    $\times $       & $\surd$             & $\times $        & $\surd$               & 80.22 \textcolor[HTML]{138906}{\small ($\uparrow 2.19$)}         & 87.35 \textcolor[HTML]{138906}{\small ($\uparrow 0.9	$)}          & 61.77 \textcolor[HTML]{138906}{\small ($\uparrow 1.15$)}          \\
    
    $\times $       & $\surd$             & $\surd$          & $\times $             & 83.32 \textcolor[HTML]{138906}{\small ($\uparrow 5.29$)}          & 90.41 \textcolor[HTML]{138906}{\small ($\uparrow 3.96$)}          & 64.89 \textcolor[HTML]{138906}{\small ($\uparrow 4.27$)}          \\
    
    $\times $       & $\surd$             & $\surd$          & $\surd$               & \textbf{84.44} \textcolor[HTML]{138906}{\small ($\uparrow 6.41$)} & \textbf{91.51} \textcolor[HTML]{138906}{\small ($\uparrow 5.06$)} & \textbf{66.17} \textcolor[HTML]{138906}{\small ($\uparrow 5.55$)}
    \\
    \bottomrule
    \end{tabular}
    
    \label{tab4_4_module_abla}
\end{table*}

\section{Experiments}
\subsection{Dataset}

We evaluate the performance of zero-shot 3D shape classification on three popular datasets {\bf ModelNet10}\cite{modelnet40}, {\bf ModelNet40}\cite{modelnet40} and {\bf ShapeNet Core55}\cite{24_chang2015shapenet}, using the complete test set without any pre-training on 3D training set.

ModelNet10 and ModelNet40 are two synthetic 3D model datasets commonly used for shape recognition and classification tasks. The former one selects 10 most common categories from ModelNet \cite{modelnet40} and provides a simplified dataset. It contains approximately 4,899 3D models from 10 different categories, with 908 3D models used for testing. In contrast, the ModelNet40 includes a greater number of categories and exhibits a richer variety of 3D shapes. It consists of 40 categories from ModelNet \cite{modelnet40} and comprises a total of 12,311 3D models, with 2,468 samples for testing. ShapeNet Core55 is a more diverse and challenging synthetic 3D dataset, widely used for 3D model analysis and understanding. It selects 55 commonly seen categories from ShapeNet \cite{24_chang2015shapenet}, comprising approximately 51,300 shapes in total, with 10,265 shapes for testing.

\subsection{Experimental Setting and Details}
Our framework is built entirely on the PyTorch and experiments are executed on the NVIDIA RTX 3090 GPU. In terms of multi-view processing, we transform the MVCNN architecture by incorporating the publicly available pre-trained visual encoders, OpenCLIP\cite{27_ilharco2openclip} and CLIP\cite{13_clip}, specifically utilizing the ViT/B-16 as the backbone network for extracting view features.
Regarding the hierarchical prompts, we employ the existing large-scale language model GPT-3.5\cite{04_brown2020language} to generate the 3D-specific prompts of candidate classes. To maintain the conciseness of the prompts, we enforce the prompt length constraint of 40.

In terms of experimental details, we designate the batch size as 4 and establish the confidence threshold parameter 
$\delta$ at 0.96. The rendering of multi-views is executed using the MV-pytorch \cite{19_hamdi2022mvtn} framework. Following the default settings of the online renderer, we uniformly set the rendering color to gray and the background color to white. The radial distance between the centroid of model and the camera is fixed at a value of 2, with an incorporation of random lighting conditions. The dimensions of the output views are set to 224×224 pixels. For the experimental results, unless explicitly stated otherwise, we concentrate on rendering views from a circular perspective. 
Except addressed particularly, we set the total number of views to 20 and the selected number of views to 4 for analysis.

\subsection{Zero-shot 3D recognition}  

In Table \ref{tab4:zero_shotacc}, we evaluate the performance of zero-shot 3D shape recognition on ModelNet10 and ModelNet40 using the complete test set without any pre-training on 3D training set.
Training-based methods utilize large-scale 3D datasets for multi-modal contrastive learning. CG3D \cite{16_hegde2023cg3d}, ULIP \cite{14_xue2023ulip} and ULIP2 \cite{15_xue2023ulip} employ an upgraded version of CLIP named SLIP \cite{28_mu2022slip}, while OpenShape \cite{18_liu2023openshape} uses the best model of OpenCLIP \cite{27_ilharco2openclip}.
On the other hand, non-training-based methods like PointCLIP V2 \cite{03_zhu2022pointclip} and DiffCLIP \cite{23_shen2023diffclip} use CLIP-ViT/B-16 \cite{13_clip}.
For MV-CLIP, we use ViT/B-16 from both CLIP \cite{13_clip} and OpenCLIP \cite{27_ilharco2openclip} for our experiments.
Besides, our approach achieves a zero-shot classification accuracy of 91.51$\%$ and 84.44$\%$ on the ModelNet10 and ModelNet40, respectively. The results are among the best in zero-shot 3D shape classification without any pre-training on 3D data, and they are comparable to the state-of-the-art results based on multi-modal contrast.

Although in the absence of 3D dataset training, our approach could still achieve competitive experimental results by fully leveraging the potential of the pre-trained model for 3D shape analysis.
Our approach selects views with clear semantics from original multiple views, which significantly enhances the discriminative power of MV-CLIP and avoids the interference of semantically ambiguous views.
Furthermore, under the first matching with hand-crafted prompts, we leverage LLMs-generated prompts that describe the characteristics within the scope of candidate labels for refined matching, effectively improving the zero-shot performance with hierarchical prompts.

\subsection{Ablation Study}

\subsubsection{Effectiveness of the designed modules.} 
As shown in Table \ref{tab4_4_module_abla}, we conduct ablation study for individual modules and different versions of CLIP. We observe that with the addition of key components and the upgrade of CLIP versions, there is a consistent improvement in zero-shot 3D shape recognition performance.

We initially analyze the benefits brought by different moudles based on the baseline that uses OpenCLIP's ViT/B-16 as the backbone. 
By incorporating only the hierarchical prompts, the accuracy improvements of $2.19\%$, $0.9\%$, and $1.15\%$ are achieved on ModelNet40, ModelNet10 and ShapeNet Core55, respectively. 
It validates the effectiveness of being tolerate with several candidates and giving a further refined matching.
By incorporating only the view selection, higher improvements are achieved, i.e., $5.29\%$, $3.96\%$, and $4.27\%$ on three datasets. 
It shows that the selected views possess clearer semantic information and effectively mitigate the adverse impact of ambiguous views.
Furthermore, it indicates that the first-layer of class-level hand-crafted prompts can effectively eliminate interfering categories, allowing the functional-level prompts of second-layer to have good candidate classes.
When both modules are employed, we achieve the best results, with improvements of $6.41\%$, $5.06\%$, and $5.55\%$ against baseline. The significant improvements demonstrates that the key modules could facilitate each other, both enhancing MV-CLIP's confidence towards 3D shape recognition.

In addition, we also conduct ablation experiments on different CLIP versions. By altering the backbone network in MV-CLIP from CLIP \cite{13_clip} to OpenCLIP \cite{27_ilharco2openclip}, we observe the gains of $16.10\%$, $15.75\%$, and $2.72\%$ on ModelNet40, ModelNet10 and ShapeNet Core55 respectively. With CLIP as backbone, the proposed modules bring the gains of $3.99\%$, $6.83\%$, and $3.8\%$.
It implies that our method has potentials to adapt to various pre-trained vision-language models and will further lift the performance of zero-shot 3D shape recognition with rising advanced pre-trained models.

\begin{table}[t]
    \centering
        \caption{View selection ablation with different number of views on ModelNet40.}
        \begin{tabular}{ccccc}
            \toprule
                      & $M$                 & $M_{selec}$   &    First-layer                 &      Second-layer                 \\ 
            \midrule
        \multirow{3}{*}{\begin{tabular}[c]{@{}c@{}}$w/ $\\ \text{Vs}\end{tabular}}   & \multirow{3}{*}{20} & 12       & 82.62          & 83.43          \\ 
                                                                            &                     & 8        & 83.21          & 84.27          \\ 
                                                                            &                     & 4        & \textbf{83.32} & \textbf{84.44} \\ 
            \midrule
        \multirow{4}{*}{\begin{tabular}[c]{@{}c@{}}$w/o$ \\ \text{Vs}\end{tabular}}  &                     &20         & 78.03          & 79.65          \\ 
                                                                            &                     &12         & 78.02          & 80.22          \\ 
                                                                            &                     &8          & 76.41          & 77.47          \\ 
                                                                            &                     &4          & 64.38          & 61.91          \\ 
            \bottomrule
        \end{tabular} 
        
        \label{tab4_4_abla_vs}

\end{table}

\begin{table}[t]
\centering
    \caption{View selection ablation with different\\aggregation types on ModelNet40.} 
    \begin{tabular}{cccc}
        \toprule
        \multirow{2}{*}{Aggregation type}                                                   & \multirow{2}{*}{Acc} \\
                                                                                             &                     \\
        \midrule
        $w/$ \text{Mean pooling}                & 66.89      \\
        \midrule
        $w/$  \text{Max pooling}              & 62.27      \\                       
        \midrule
        $w/o$ \text{Pooling}                 & \textbf{84.44}        \\
        \bottomrule                                
    \end{tabular}
    
\label{tab4_4:agr}
\end{table} 
\begin{table}[t]
        \centering
        \caption{View selection ablation with different configurations of rendering on ModelNet40.}
        \begin{tabular}{cccc}
            \toprule
            \multicolumn{2}{c}{View configuration}  &First-layer   & Second-layer    \\ 
            \midrule
            \multirow{2}{*}{Random}      & $w/o \;  \text{Vs}$   & 57.90             & 58.02        \\
                                         & $w/ \;  \text{Vs} $   &75.08 \textcolor[HTML]{138906}{\small ($\uparrow 17.18$)}        & 75.72 \textcolor[HTML]{138906}{\small ($\uparrow 17.7$)}       \\
            \midrule                             
            \multirow{2}{*}{Spherical}   & $w/o \; \text{Vs}$   & 64.02            & 61.79            \\
                                         & $w/ \;  \text{Vs} $   &69.20 \textcolor[HTML]{138906}{\small ($\uparrow 5.18$)}              & 69.43 \textcolor[HTML]{138906}{\small ($\uparrow 7.64$)}       \\
            \midrule
            \multirow{2}{*}{Circular}    & $w/o \; \text{Vs}$   &78.03             & 80.22              \\
                                         & $w/ \;  \text{Vs} $   &\textbf{83.32} \textcolor[HTML]{138906}{\small ($\uparrow 5.29$)}             & \textbf{84.44} \textcolor[HTML]{138906}{\small ($\uparrow 4.22$)}       \\
            \bottomrule
        \end{tabular}

        \label{tab4_4_abla_vc}

\end{table}

\begin{table}[t]
\centering
    \caption{Average accuracy for individual view on\\ModelNet40 (only first layer involved).} 
    \begin{tabular}{ccc}
    \toprule
    \multirow{2}{*}{Dataset}                                           & \multirow{2}{*}{\begin{tabular}[c]{@{}c@{}}All views\\ (20)\end{tabular}}            & \multirow{2}{*}{\begin{tabular}[c]{@{}c@{}}Selected views\\ (4)\end{tabular}} \\
                                                                         &                     &                     \\
    \midrule
    MN40                                  &57.54              & 79.97 \\
    \midrule
    MN10                                  & 66.37             & 86.71 \\
    \midrule                                                                           
    SN55                             & 51.23             & 64.07 \\
    \bottomrule
    \end{tabular}
    \label{tab:apa}
\end{table} 

\begin{figure}[t!]
    \centering
    \includegraphics[width=0.8\linewidth]{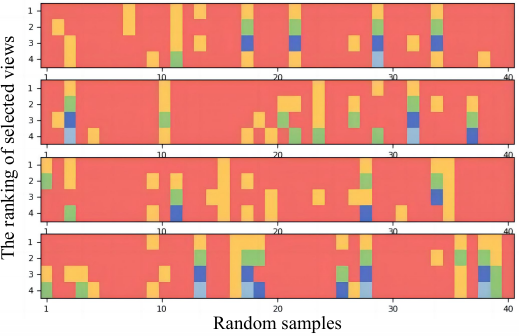}
    \caption{The decisions of the selected 4 views of $40\times4$ randomly chosen samples on ModelNet40.
}
    \label{fig:01_sample}
\end{figure}

\subsubsection{Discussions on the view selection module.} 
Table \ref{tab4_4_abla_vs}, \ref{tab4_4_abla_vc} and \ref{tab4_4:agr} show the classification results on ModelNet40 for different number of views, view rendering configurations and multi-view aggregation types, respectively. 
Additionally, we show the superiority of selected views with the average accuracy of individual view image and further discuss the variance of selected views, as shown in Table \ref{tab:apa} and Figure~\ref{fig:01_sample}.
\paragraph{ Different number of views.} In Table \ref{tab4_4_abla_vs}, we find that in the case of multi-view setting without selection, once the number of views exceeds 12, the performance of 3D shape classification improves slowly (almost no gains).  
With the adoption of view selection, the zero-shot classification performance is steadily improved, but selecting too many views will bring redundant information and slightly hinder the performance. 
\paragraph{ Different configurations of rendering.} Table \ref{tab4_4_abla_vc} shows not only the importance of view quality, but also the effectiveness of view selection to filter low-quality views. Furthermore, the gains caused by view selection will increase when the view sets contain poorer viewpoints. Please note that the camera positions of different configurations are shown in the supplement and Figure~\ref{fig:06_vision} gives an example of captured views under these configurations.
\paragraph{ Different multi-view aggregation types.} In Table \ref{tab4_4:agr}, given that the backbone network of MV-CLIP is a 2D pre-trained visual encoder, aggregating the features of multiple views negatively affects the alignment between views and texts. Therefore, unlike MVCNN, MV-CLIP fuses the prediction instead of any pooling of features.
\paragraph{ Average accuracy for individual view.} Table \ref{tab:apa} illustrates the average view prediction accuracy of the selected views and the total views on three datasets. We find that the views within the selected set have much higher accuracy than the rest, which shows that view selection choose views with high prediction confidence from multiple views, and filters the negative influence of views with ambiguous semantics on the final decision.

\paragraph{ Variance within the selected views.} Figure~\ref{fig:01_sample} shows the variance within the selected view decisions, where red patch indicates the right decision and different colors in each column show the disagreement. We find that the views selected in most samples have similar decisions, and we regard the similar decision as a kind of weighting where the fused decision of selected views is impacted by majority voting. 
Furthermore, we tried another variant of view selection by keeping more diversity in decisions of selected views. Along the increase of entropy, we select 4 views with different decisions, but the accuracy only reaches 74.68\% (84.44\% for the proposed method). It indicates that incorporating views with contrary decisions may introduce ambiguity into the final decision-making process.

\subsubsection{Discussions on the hierarchical prompts module.} 
\paragraph{Different prompt settings.} Several variants for prompting have been tried, with emphasis on only visual characteristics, only functional features, fusion of the both, difference between candidates and a combination of visual and functional features. Table~\ref{tab:hp_ab} illustrates the results of different variants, which shows that both visual and functional characteristics are important and each aspect contributes comparatively. Putting emphasis on the difference between candidates is also an alternative way for the second matching. Comparative experiments with templates from DiffCLIP~\cite{23_shen2023diffclip} are shown in Table~\ref{tab:comparative_templates}, demonstrating the superior performance of the proposed template. By comparing baseline (w/o view selection) and first-layer in Table~\ref{tab:comparative_templates}, we find that prompts affect the performance, but the view selection process is not quite sensitive to the initial prompt design (i.e., with comparative benefits).

\begin{figure*}[t]
    \centering
    \begin{minipage}[b]{0.3\textwidth}
        \includegraphics[width=\textwidth]{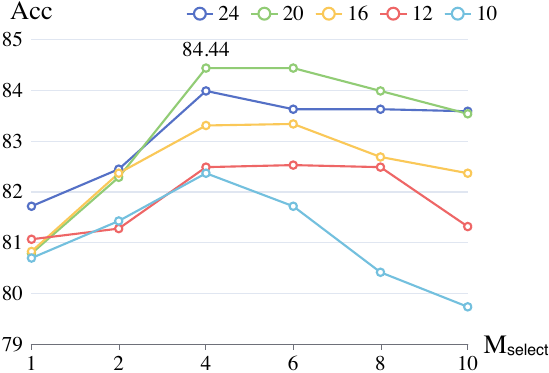}
        \subcaption[]{\fontsize{8}{10}\selectfont View selection}
        \label{fig:sensi_Msel}
    \end{minipage}
    \hfill
    \begin{minipage}[b]{0.3\textwidth}
        \includegraphics[width=\textwidth]{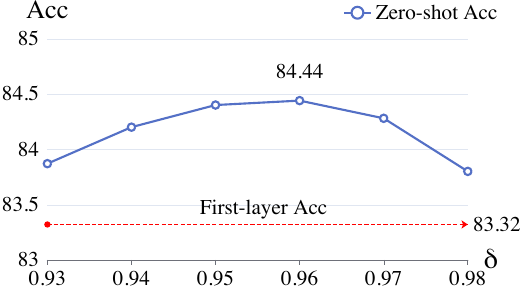}
        \subcaption[]{\fontsize{8}{10}\selectfont Confidence threshold $\delta$}
        \label{fig:sensi_thre}
    \end{minipage}
    \hfill
    \begin{minipage}[b]{0.3\textwidth}
        \includegraphics[width=\textwidth]{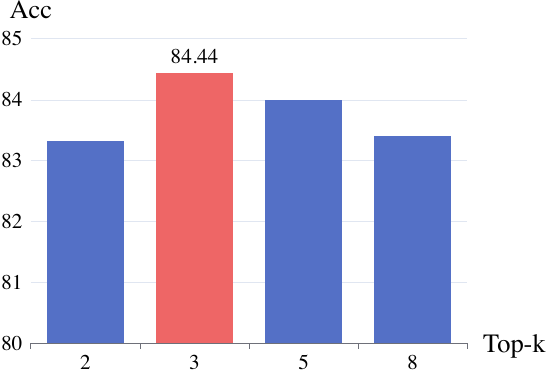}
        \subcaption[]{\fontsize{8}{10}\selectfont Top-k classes}
        \label{fig:sensi_topk}
    \end{minipage}
    \caption{Sensitivity analysis of view selection and hierarchical prompts on ModelNet40. }
    \label{fig:04_sensi}
    
\end{figure*}

\begin{table}[t!]\scriptsize
    \centering
    \caption{Hierarchical prompts ablation with different prompt settings on ModelNet40. All results are based on an accuracy of 83.32\% at the first layer.} 
    \begin{tabular}{cc}
    \toprule
    Prompts setting                                                                     & Acc \\
    \midrule
    Only visual characteristics prompts of candidate classes                            & 83.72    \\
    \midrule
    Only functional features prompts of candidate classes                               & 83.75    \\
    \midrule
    Fusion of visual characteristics and functional features prompts               &  84.31 \\
    \midrule
    The prompts of difference between candidate classes                                 &  84.42   \\
    \midrule
    The visual characteristics and functional features of candidate classes             &\textbf{84.44}  \\
    \bottomrule
    \end{tabular}
    \label{tab:hp_ab}
\end{table}

\begin{figure}[t!]
    \centering
    \includegraphics[width=1\linewidth]{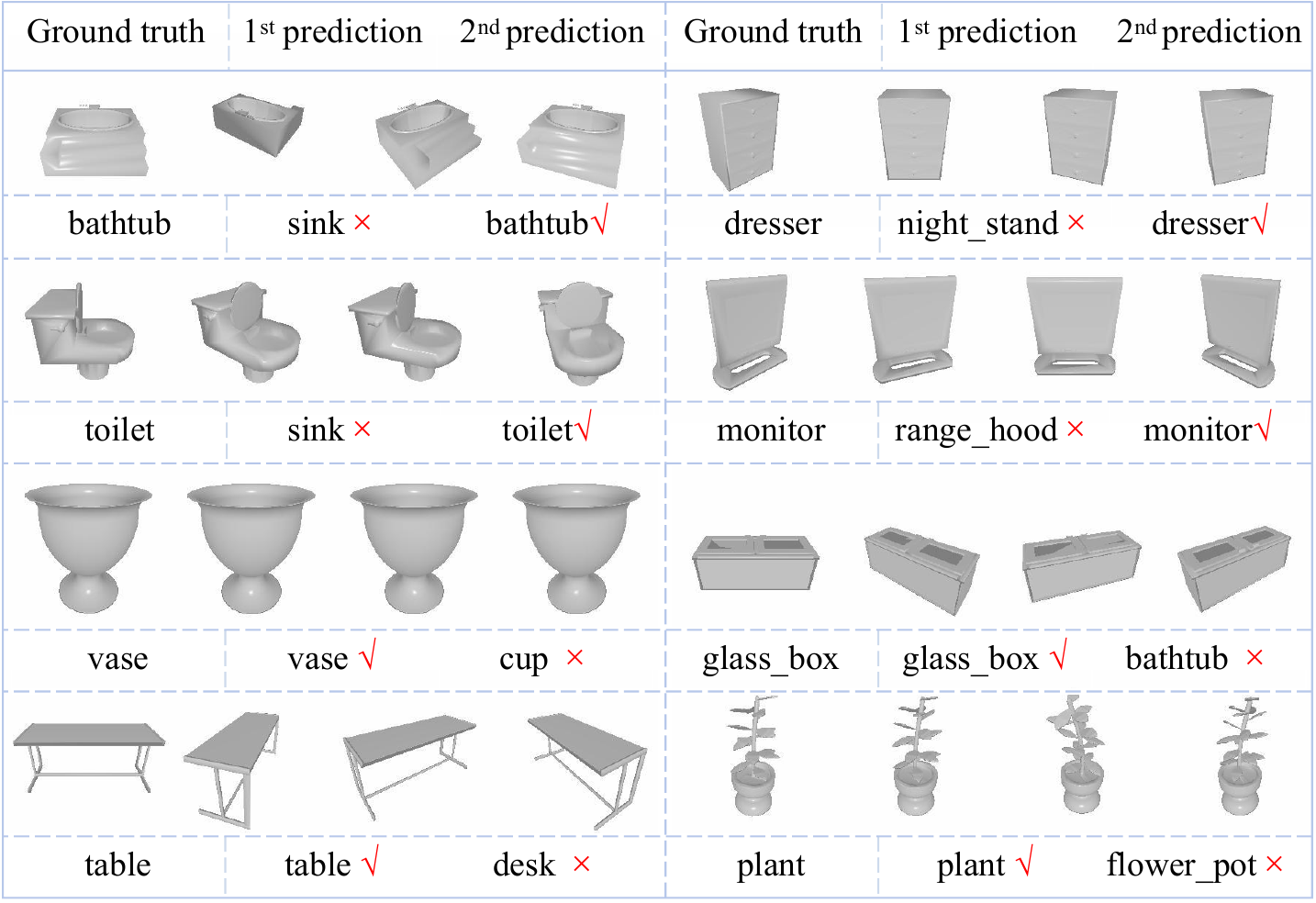}
    \caption{ Visualization of successful and failed cases for second matching. Note that the text below the selected views represents the true label, the first-layer prediction, and the second-layer prediction, respectively. 
    }
    \label{fig:06_hp_crcw}
\end{figure}

\paragraph{The correction capability of hierarchical prompts.} 
As shown in Figure~\ref{fig:06_hp_crcw}, we visualize some successful and failed cases caused by the second-layer matching based on hierarchical prompts. Sometimes the second matching changes the decision of first-layer matching. Statistically, the number of successful corrections by second-layer matching is approximately 2.3 times that of failure cases.  Additionally, we find that in most of the successful cases, the unique function and shape characteristics compared to other categories are captured in the second match. For example, bathtub is used for bathing, dresser has multiple drawers for storing items, toilet has a water tank, and monitor has a rectangular screen. We also summarize the failed reasons as: (1) Visual similarity, e.g., vase and cup. (2) Limited rendering quality, e.g., views of glass\_box do not display transparency which is an important attribute in the second layer prompt. (3) Prompt quality, e.g., desk has more explicit descriptions than table. (4) Co-existence of multiple objects, e.g., flower\_pot and plant.

\subsection{Sensitivity Analysis}

We discuss the prediction accuracy affected by different views settings of view selection, different thresholds $\delta$ and different numbers of candidate classes in hierarchical prompts, as shown in Figure~\ref{fig:04_sensi}.
According to our observations, selecting too few views results in insufficient information, while selecting too many views reduces the benefits of view selection under a defined total number of views, which are not desirable for zero-shot 3D recognition. 
Furthermore, if the total number of views is excessively high, it can lead to high similarity among different views, resulting in the selected views being similar to each other and lack of diversity.

As shown in Figure~\ref{fig:sensi_thre}, a lower threshold results in more high-confidence samples and less reliance on hierarchical matching, thus the accuracy tends to be closer to results without using hierarchical prompts. Conversely, a higher threshold increases the number of low-confidence samples.
It implies that for high-confident samples, the first-layer prompts work well and the improvement introduced by hierarchical prompts is limited. 
To balance accuracy with efficiency in hierarchical prompts matching, we set the threshold at 0.96 to refine the classification of unreliable samples. 

Additionally, we test on the classification results under the different numbers of candidate classes, as shown in Figure~\ref{fig:sensi_topk}.
We can observe that when the number of candidate labels equals to 3, it leads to the maximum gain in hierarchical prompts.
It reflects that the LLMs powered prompts in the second layer, which describe fine-grained characteristics of categories, performs better within a limited number of classes.

\begin{table}[t!]\scriptsize
    \centering
    \caption{Comparative experiments with DiffCLIP templates.} 
    \begin{tabular}{cccc}
    \toprule
    Prompt templates for first-layer                       &w/o Vs         & First-layer           & Second-layer \\
    \midrule
    a 3D rendered image of [class].                        & 68.43        &  76.80                &78.21\\
    \midrule
    {[class]} with white background.                        & 72.48       &  79.46                &80.47\\
    \midrule
    \multirow{2}{*}{\begin{tabular}[c]{@{}c@{}}a 3D rendered image of \\{[class]} with white background. \end{tabular}}   & \multirow{2}{*}{74.43}         & \multirow{2}{*}{81.32}                & \multirow{2}{*}{81.87}\\
                                                                            &   &   & \\
    
    \midrule
    a photo of a [class].                                  & 77.07        & 81.87                 &82.90  \\
    \midrule
    a rendered image of [class].                         & 77.26       & 82.40                 &83.14\\
    
    \midrule
    \multirow{2}{*}{\begin{tabular}[c]{@{}c@{}}a synthetic 3D model view of  \\{[class]} with different angles. \end{tabular}}   & \multirow{2}{*}{\textbf{78.03} }        & \multirow{2}{*}{\textbf{83.32} }                & \multirow{2}{*}{\textbf{84.44}}\\
                                                                            &   &   & \\
    \bottomrule
    \end{tabular}
    \label{tab:comparative_templates}
\end{table}

\subsection{Experiments with Offline Rendering}

\begin{figure}[t]
    \centering
    \includegraphics[width=0.9\linewidth]{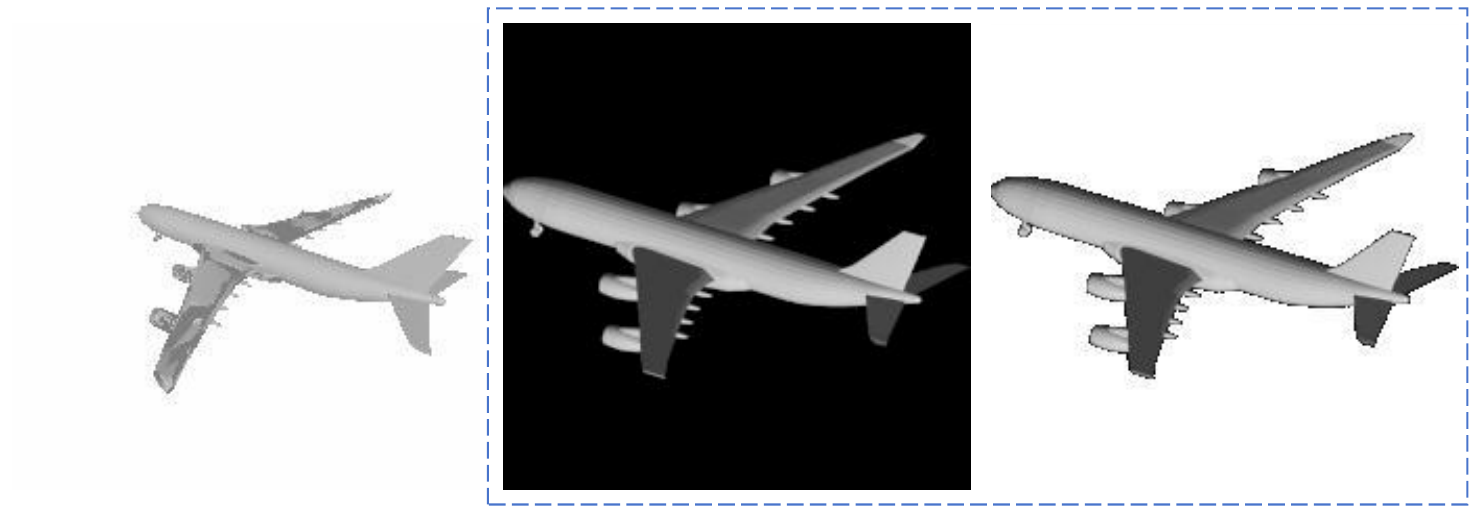}
    \caption{Comparison between online and offline rendered views. Left: an example of online rendering with sub-optimal rendering quality. Middle and right: an example of offline rendering with relatively better rendering quality.}
    \label{fig:offline}
\end{figure}

In this section, we present the experimental results obtained using offline views \cite{offline_su2018deeper}. As depicted in Figure~\ref{fig:offline}, views rendered offline exhibit higher quality with better shadow, but it suffers from slower rendering speed and storage requirement for 3D shapes. In contrast, online rendering views allow for timely operation and can cooperate with view selection to save storage. Therefore, we employ an online renderer \cite{online_hamdi2022mvtn} in our method and consider the results from the online renderer as the final outcome.

For consistency in experimental comparison, we alter the background color of the offline rendered views to white. Based on the experimental results shown in Table~\ref{tab:offline}, we observe that offline rendered views perform better than online ones. Furthermore, under offline view rendering, our approach achieves a consistent $4.83\%$ improvement, which substantiates the effectiveness of the modules we propose.

\begin{table}[t!]
    \centering
    \caption{ Zero-shot results on ModelNet40 (12 views) with different renderers. With a better renderer, the proposed modules still facilitate the performance.}
    \begin{tabular}{ccccc}
    \toprule
    \multirow{2}{*}{Renderer} & \multirow{2}{*}{Method}               & \multirow{2}{*}{$M$}     &\multirow{2}{*}{$M_{selec}$}    & \multirow{2}{*}{Acc} \\ 
                              &                                       &                             &                                   &                      \\  
    \midrule
    \multirow{5}{*}{Offline}  & \multirow{2}{*}{Ours}                 & \multirow{2}{*}{12}    & 6          & 85.05                \\
                              &                                       &                        & 4          & \textbf{85.17}       \\ \cline{2-5}
                              & \multirow{3}{*}{w/o designed modules} &                        & 12          & 80.34                \\
                              &                                       &                        & 6          & 76.78                \\
                              &                                       &                        & 4          & 64.16                \\
    \midrule
    \multirow{5}{*}{Online}   & \multirow{2}{*}{Ours}                 & \multirow{2}{*}{12}    & 6          & 82.53                \\
                              &                                       &                        & 4          & 82.49                \\ \cline{2-5}
                              & \multirow{3}{*}{w/o designed modules} &                        & 12          & 78.42                \\
                              &                                       &                        & 6          & 75.04                \\
                              &                                       &                        & 4          & 64.38                \\
    \bottomrule
    \end{tabular}
    \label{tab:offline}
\end{table}

\subsection{Visualization}
\begin{figure}[t!]
	\centering
 	\includegraphics[width=0.48\textwidth]{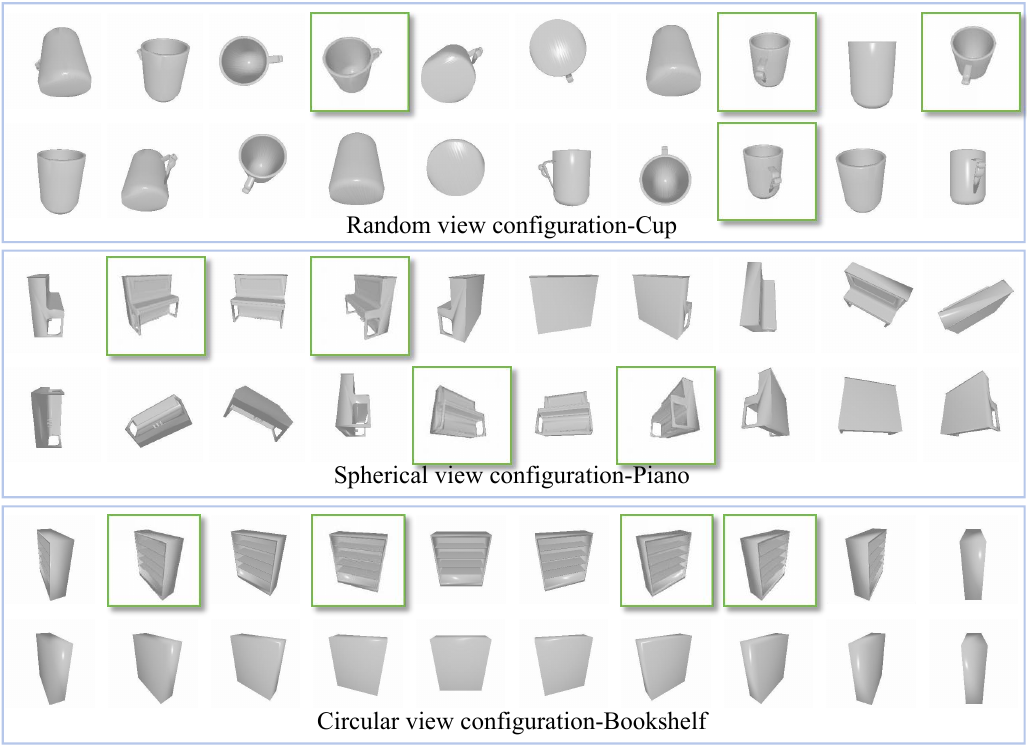}
	\caption{Visualization of multiple views from the models in ModelNet40. Note that the selected views are indicated by green boxes.}
	\label{fig:06_vision}       
\end{figure}

As shown in Figure~\ref{fig:06_vision}, we visualize a subset of the selected views. It is observable that the chosen views typically encompass more comprehensive category features and have relatively clearer semantic information. In contrast, views with fewer features or even lacking semantic contents, like the bottom of a cup, the back of a piano or the side of a bookshelf, are not selected. Consequently, the view selection based on entropy minimization effectively reduces the redundancy present in the rendered multiple views of a 3D shape.

\begin{figure}[t!]
    \centering
    \includegraphics[width=1\linewidth]{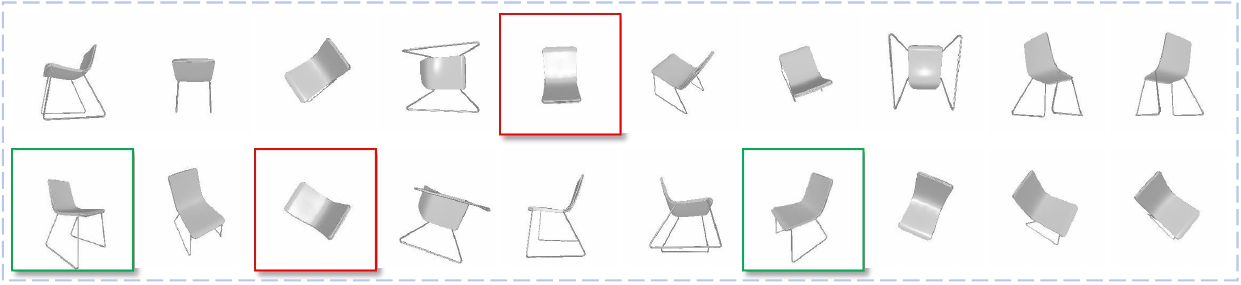}
    \caption{An example of incorrect view selection. The selected views exhibit semantic inconsistency and do not align with the ground truth.}
    \label{fig:limitation}
\end{figure}
\section{Limitation}

The proposed MV-CLIP achieves state-of-the-art zero-shot 3D recognition performance using pre-trained vision-language models without any fine-tuning with 3D data, comparable to the most advanced technologies in the field. However, further exploration is needed in the following aspects. 
Firstly, some predefined settings such as the number of selected views and the number of candidate labels could be adaptive. 
Secondly, 
the selected views lack semantic consistency constraints. As illustrated in Figure~\ref{fig:limitation}, the views outlined in red are semantically identified as sink and laptop, which are inconsistent with the other views, i.e., chair.
Finally, it is worth mentioning that GPT-generated prompts for second layer can be performed offline, so there is no heavy reliance on API accessibility. For encountering objects that are out of
the scope of prepared classes, we need online generation. Under
this situation, alternative strategy avoiding API dependency could
be further explored. For example, we can discover approximate representation for the second layer with the
help of its neighbouring classes computed from the first layer.

\section{Conclusion}
We design a zero-shot 3D recognition pipeline based on MV-CLIP to fully leverage the large-scale pre-trained models.
We utilize the pre-trained visual encoder to evaluate the prediction confidence of the rendered multiple views, and explicitly select views with clearer semantics. 
In addition, we combine hand-crafted prompts with 3D-specific prompts powered by LLMs to form hierarchical prompts, which refine the first-layer prediction and achieve more accurate zero-shot performance via the second-layer matching.
The experimental results demonstrate that we obtain superior performance of zero-shot 3D shape recognition by directly utilizing pre-trained models without any pre-training on 3D dataset, and this paper also discovers some interesting findings in prompt engineering.

\bibliographystyle{ieeetr}
\bibliography{main}

\begin{thebibliography}{10}

\bibitem{tcsvt_li2023focus}
T.-B. Li, A.-A. Liu, D.~Song, W.-H. Li, X.-Y. Li, and Y.-T. Su, ``Focus on hard samples: Hierarchical unbiased constraints for cross-domain 3d model retrieval,'' {\em IEEE Transactions on Circuits and Systems for Video Technology}, vol.~33, no.~11, pp.~7036--7049, 2023.

\bibitem{tcsvt_su2019joint}
Y.~Su, Y.~Li, W.~Nie, D.~Song, and A.-A. Liu, ``Joint heterogeneous feature learning and distribution alignment for 2d image-based 3d object retrieval,'' {\em IEEE Transactions on Circuits and Systems for Video Technology}, vol.~30, no.~10, pp.~3765--3776, 2019.

\bibitem{tcsvt_sun2021joint}
J.~Sun, Z.~Wang, W.~Wang, H.~Li, F.~Sun, and Z.~Ding, ``Joint adaptive dual graph and feature selection for domain adaptation,'' {\em IEEE Transactions on Circuits and Systems for Video Technology}, vol.~32, no.~3, pp.~1453--1466, 2021.

\bibitem{3D_zero1}
A.~Cheraghian, S.~Rahman, and L.~Petersson, ``Zero-shot learning of 3d point cloud objects,'' in {\em 2019 16th International Conference on Machine Vision Applications (MVA)}, pp.~1--6, IEEE, 2019.

\bibitem{3D_zero2}
A.~Cheraghian, S.~Rahman, T.~F. Chowdhury, D.~Campbell, and L.~Petersson, ``Zero-shot learning on 3d point cloud objects and beyond,'' {\em International Journal of Computer Vision}, vol.~130, no.~10, pp.~2364--2384, 2022.

\bibitem{3D_zero3}
Y.~Su, J.~Li, W.~Li, Z.~Gao, H.~Chen, X.~Li, and A.-A. Liu, ``Semantically guided projection for zero-shot 3d model classification and retrieval,'' {\em Multimedia Systems}, vol.~28, no.~6, pp.~2437--2451, 2022.

\bibitem{13_clip}
A.~Radford, J.~W. Kim, C.~Hallacy, A.~Ramesh, G.~Goh, S.~Agarwal, G.~Sastry, A.~Askell, P.~Mishkin, J.~Clark, {\em et~al.}, ``Learning transferable visual models from natural language supervision,'' in {\em International conference on machine learning}, pp.~8748--8763, PMLR, 2021.

\bibitem{14_xue2023ulip}
L.~Xue, M.~Gao, C.~Xing, R.~Mart{\'\i}n-Mart{\'\i}n, J.~Wu, C.~Xiong, R.~Xu, J.~C. Niebles, and S.~Savarese, ``Ulip: Learning a unified representation of language, images, and point clouds for 3d understanding,'' in {\em Proceedings of the IEEE/CVF Conference on Computer Vision and Pattern Recognition}, pp.~1179--1189, 2023.

\bibitem{16_hegde2023cg3d}
D.~Hegde, J.~M.~J. Valanarasu, and V.~Patel, ``Clip goes 3d: Leveraging prompt tuning for language grounded 3d recognition,'' in {\em Proceedings of the IEEE/CVF International Conference on Computer Vision}, pp.~2028--2038, 2023.

\bibitem{01_zhang2022pointclip}
R.~Zhang, Z.~Guo, W.~Zhang, K.~Li, X.~Miao, B.~Cui, Y.~Qiao, P.~Gao, and H.~Li, ``Pointclip: Point cloud understanding by {CLIP},'' in {\em {IEEE/CVF} Conference on Computer Vision and Pattern Recognition, {CVPR} 2022, New Orleans, LA, USA, June 18-24, 2022}, pp.~8542--8552, {IEEE}, 2022.

\bibitem{03_zhu2022pointclip}
X.~Zhu, R.~Zhang, B.~He, Z.~Guo, Z.~Zeng, Z.~Qin, S.~Zhang, and P.~Gao, ``Pointclip v2: Prompting clip and gpt for powerful 3d open-world learning,'' pp.~2639--2650, 2023.

\bibitem{23_shen2023diffclip}
S.~Shen, Z.~Zhu, L.~Fan, H.~Zhang, and X.~Wu, ``Diffclip: Leveraging stable diffusion for language grounded 3d classification,'' {\em arXiv preprint arXiv:2305.15957}, 2023.

\bibitem{02_huang2023clip2point}
T.~Huang, B.~Dong, Y.~Yang, X.~Huang, R.~W. Lau, W.~Ouyang, and W.~Zuo, ``Clip2point: Transfer clip to point cloud classification with image-depth pre-training,'' in {\em Proceedings of the IEEE/CVF International Conference on Computer Vision}, pp.~22157--22167, 2023.

\bibitem{04_brown2020language}
T.~Brown, B.~Mann, N.~Ryder, M.~Subbiah, J.~D. Kaplan, P.~Dhariwal, A.~Neelakantan, P.~Shyam, G.~Sastry, A.~Askell, {\em et~al.}, ``Language models are few-shot learners,'' {\em Advances in neural information processing systems}, vol.~33, pp.~1877--1901, 2020.

\bibitem{29_wang2023beyond}
H.~Wang, J.~Tang, J.~Ji, X.~Sun, R.~Zhang, Y.~Ma, M.~Zhao, L.~Li, Z.~Zhao, T.~Lv, {\em et~al.}, ``Beyond first impressions: Integrating joint multi-modal cues for comprehensive 3d representation,'' in {\em Proceedings of the 31st ACM International Conference on Multimedia}, pp.~3403--3414, 2023.

\bibitem{30_sanghi2023sketch}
A.~Sanghi, P.~K. Jayaraman, A.~Rampini, J.~G. Lambourne, H.~Shayani, E.~Atherton, and S.~A. Taghanaki, ``Sketch-a-shape: Zero-shot sketch-to-3d shape generation,'' {\em CoRR}, vol.~abs/2307.03869, 2023.

\bibitem{31_zhang2023clip}
J.~Zhang, R.~Dong, and K.~Ma, ``Clip-fo3d: Learning free open-world 3d scene representations from 2d dense clip,'' {\em arXiv preprint arXiv:2303.04748}, 2023.

\bibitem{17_zeng2023clip2}
Y.~Zeng, C.~Jiang, J.~Mao, J.~Han, C.~Ye, Q.~Huang, D.-Y. Yeung, Z.~Yang, X.~Liang, and H.~Xu, ``Clip2: Contrastive language-image-point pretraining from real-world point cloud data,'' in {\em Proceedings of the IEEE/CVF Conference on Computer Vision and Pattern Recognition}, pp.~15244--15253, 2023.

\bibitem{18_liu2023openshape}
M.~Liu, R.~Shi, K.~Kuang, Y.~Zhu, X.~Li, S.~Han, H.~Cai, F.~Porikli, and H.~Su, ``Openshape: Scaling up 3d shape representation towards open-world understanding,'' {\em arXiv preprint arXiv:2305.10764}, 2023.

\bibitem{15_xue2023ulip}
L.~Xue, N.~Yu, S.~Zhang, J.~Li, R.~Mart{\'\i}n-Mart{\'\i}n, J.~Wu, C.~Xiong, R.~Xu, J.~C. Niebles, and S.~Savarese, ``Ulip-2: Towards scalable multimodal pre-training for 3d understanding,'' {\em arXiv preprint arXiv:2305.08275}, 2023.

\bibitem{tcsvtr_huang2021learning}
J.~Huang, W.~Yan, G.~Li, T.~Li, and S.~Liu, ``Learning disentangled representation for multi-view 3d object recognition,'' {\em IEEE Transactions on Circuits and Systems for Video Technology}, vol.~32, no.~2, pp.~646--659, 2021.

\bibitem{08_wei2020view}
X.~Wei, R.~Yu, and J.~Sun, ``View-gcn: View-based graph convolutional network for 3d shape analysis,'' in {\em Proceedings of the IEEE/CVF Conference on Computer Vision and Pattern Recognition}, pp.~1850--1859, 2020.

\bibitem{05_bradski1994recognition}
G.~Bradski and S.~Grossberg, ``Recognition of 3-d objects from multiple 2-d views by a self-organizing neural architecture,'' in {\em From Statistics to Neural Networks: Theory and Pattern Recognition Applications}, pp.~349--375, Springer, 1994.

\bibitem{06_su2015multi}
H.~Su, S.~Maji, E.~Kalogerakis, and E.~Learned-Miller, ``Multi-view convolutional neural networks for 3d shape recognition,'' in {\em Proceedings of the IEEE international conference on computer vision}, pp.~945--953, 2015.

\bibitem{09_hamdi2021mvtn}
A.~Hamdi, S.~Giancola, and B.~Ghanem, ``Mvtn: Multi-view transformation network for 3d shape recognition,'' in {\em Proceedings of the IEEE/CVF International Conference on Computer Vision}, pp.~1--11, 2021.

\bibitem{10_goyal2021revisiting}
A.~Goyal, H.~Law, B.~Liu, A.~Newell, and J.~Deng, ``Revisiting point cloud shape classification with a simple and effective baseline,'' in {\em International Conference on Machine Learning}, pp.~3809--3820, PMLR, 2021.

\bibitem{11_zhang2022point}
R.~Zhang, Z.~Guo, P.~Gao, R.~Fang, B.~Zhao, D.~Wang, Y.~Qiao, and H.~Li, ``Point-m2ae: multi-scale masked autoencoders for hierarchical point cloud pre-training,'' {\em Advances in neural information processing systems}, vol.~35, pp.~27061--27074, 2022.

\bibitem{12_zhang2023learning}
R.~Zhang, L.~Wang, Y.~Qiao, P.~Gao, and H.~Li, ``Learning 3d representations from 2d pre-trained models via image-to-point masked autoencoders,'' in {\em Proceedings of the IEEE/CVF Conference on Computer Vision and Pattern Recognition}, pp.~21769--21780, 2023.

\bibitem{37_liu2023pre}
P.~Liu, W.~Yuan, J.~Fu, Z.~Jiang, H.~Hayashi, and G.~Neubig, ``Pre-train, prompt, and predict: A systematic survey of prompting methods in natural language processing,'' {\em ACM Computing Surveys}, vol.~55, no.~9, pp.~1--35, 2023.

\bibitem{38_jiang2020can}
Z.~Jiang, F.~F. Xu, J.~Araki, and G.~Neubig, ``How can we know what language models know?,'' {\em Transactions of the Association for Computational Linguistics}, vol.~8, pp.~423--438, 2020.

\bibitem{39_wallace2019universal}
E.~Wallace, S.~Feng, N.~Kandpal, M.~Gardner, and S.~Singh, ``Universal adversarial triggers for attacking and analyzing nlp,'' {\em arXiv preprint arXiv:1908.07125}, 2019.

\bibitem{36_devlin2018bert}
J.~Devlin, M.-W. Chang, K.~Lee, and K.~Toutanova, ``Bert: Pre-training of deep bidirectional transformers for language understanding,'' {\em arXiv preprint arXiv:1810.04805}, 2018.

\bibitem{41_zhou2022conditional}
K.~Zhou, J.~Yang, C.~C. Loy, and Z.~Liu, ``Conditional prompt learning for vision-language models,'' in {\em Proceedings of the IEEE/CVF Conference on Computer Vision and Pattern Recognition}, pp.~16816--16825, 2022.

\bibitem{42_zhou2022learning}
K.~Zhou, J.~Yang, C.~C. Loy, and Z.~Liu, ``Learning to prompt for vision-language models,'' {\em International Journal of Computer Vision}, vol.~130, no.~9, pp.~2337--2348, 2022.

\bibitem{43_khattak2023maple}
M.~U. Khattak, H.~Rasheed, M.~Maaz, S.~Khan, and F.~S. Khan, ``Maple: Multi-modal prompt learning,'' in {\em Proceedings of the IEEE/CVF Conference on Computer Vision and Pattern Recognition}, pp.~19113--19122, 2023.

\bibitem{44_jia2022visual}
M.~Jia, L.~Tang, B.-C. Chen, C.~Cardie, S.~Belongie, B.~Hariharan, and S.-N. Lim, ``Visual prompt tuning,'' in {\em European Conference on Computer Vision}, pp.~709--727, Springer, 2022.

\bibitem{45_bahng2022visual}
H.~Bahng, A.~Jahanian, S.~Sankaranarayanan, and P.~Isola, ``Visual prompting: Modifying pixel space to adapt pre-trained models,'' {\em arXiv preprint arXiv:2203.17274}, vol.~3, pp.~11--12, 2022.

\bibitem{46_guo2023viewrefer}
Z.~Guo, Y.~Tang, R.~Zhang, D.~Wang, Z.~Wang, B.~Zhao, and X.~Li, ``Viewrefer: Grasp the multi-view knowledge for 3d visual grounding,'' pp.~15372--15383, 2023.

\bibitem{48_pratt2023does}
S.~Pratt, I.~Covert, R.~Liu, and A.~Farhadi, ``What does a platypus look like? generating customized prompts for zero-shot image classification,'' in {\em Proceedings of the IEEE/CVF International Conference on Computer Vision}, pp.~15691--15701, 2023.

\bibitem{47_zhang2023prompt}
R.~Zhang, X.~Hu, B.~Li, S.~Huang, H.~Deng, Y.~Qiao, P.~Gao, and H.~Li, ``Prompt, generate, then cache: Cascade of foundation models makes strong few-shot learners,'' in {\em Proceedings of the IEEE/CVF Conference on Computer Vision and Pattern Recognition}, pp.~15211--15222, 2023.

\bibitem{49_novack2023chils}
Z.~Novack, J.~McAuley, Z.~C. Lipton, and S.~Garg, ``Chils: Zero-shot image classification with hierarchical label sets,'' in {\em International Conference on Machine Learning}, pp.~26342--26362, PMLR, 2023.

\bibitem{19_hamdi2022mvtn}
A.~Hamdi, F.~AlZahrani, S.~Giancola, and B.~Ghanem, ``Mvtn: Learning multi-view transformations for 3d understanding,'' 2022.

\bibitem{07_kanezaki2018rotationnet}
A.~Kanezaki, Y.~Matsushita, and Y.~Nishida, ``Rotationnet: Joint object categorization and pose estimation using multiviews from unsupervised viewpoints,'' in {\em Proceedings of the IEEE conference on computer vision and pattern recognition}, pp.~5010--5019, 2018.

\bibitem{20_zhao2021point}
H.~Zhao, L.~Jiang, J.~Jia, P.~H. Torr, and V.~Koltun, ``Point transformer,'' in {\em Proceedings of the IEEE/CVF international conference on computer vision}, pp.~16259--16268, 2021.

\bibitem{28_mu2022slip}
N.~Mu, A.~Kirillov, D.~Wagner, and S.~Xie, ``Slip: Self-supervision meets language-image pre-training,'' in {\em European Conference on Computer Vision}, pp.~529--544, Springer, 2022.

\bibitem{24_chang2015shapenet}
A.~X. Chang, T.~A. Funkhouser, L.~J. Guibas, P.~Hanrahan, Q.~Huang, Z.~Li, S.~Savarese, M.~Savva, S.~Song, H.~Su, J.~Xiao, L.~Yi, and F.~Yu, ``Shapenet: An information-rich 3d model repository,'' {\em CoRR}, vol.~abs/1512.03012, 2015.

\bibitem{21_yu2022point}
X.~Yu, L.~Tang, Y.~Rao, T.~Huang, J.~Zhou, and J.~Lu, ``Point-bert: Pre-training 3d point cloud transformers with masked point modeling,'' in {\em Proceedings of the IEEE/CVF Conference on Computer Vision and Pattern Recognition}, pp.~19313--19322, 2022.

\bibitem{25_deitke2023objaverse}
M.~Deitke, D.~Schwenk, J.~Salvador, L.~Weihs, O.~Michel, E.~VanderBilt, L.~Schmidt, K.~Ehsani, A.~Kembhavi, and A.~Farhadi, ``Objaverse: A universe of annotated 3d objects,'' in {\em Proceedings of the IEEE/CVF Conference on Computer Vision and Pattern Recognition}, pp.~13142--13153, 2023.

\bibitem{27_ilharco2openclip}
G.~Ilharco, M.~Wortsman, R.~Wightman, C.~Gordon, N.~Carlini, R.~Taori, A.~Dave, V.~Shankar, H.~Namkoong, J.~Miller, {\em et~al.}, ``Openclip, july 2021,'' {\em If you use this software, please cite it as below}, vol.~2, no.~4, p.~5.

\bibitem{22_qi2023contrast}
Z.~Qi, R.~Dong, G.~Fan, Z.~Ge, X.~Zhang, K.~Ma, and L.~Yi, ``Contrast with reconstruct: Contrastive 3d representation learning guided by generative pretraining,'' {\em arXiv preprint arXiv:2302.02318}, 2023.

\bibitem{modelnet40}
Z.~Wu, S.~Song, A.~Khosla, F.~Yu, L.~Zhang, X.~Tang, and J.~Xiao, ``3d shapenets: A deep representation for volumetric shapes,'' in {\em Proceedings of the IEEE conference on computer vision and pattern recognition}, pp.~1912--1920, 2015.

\bibitem{offline_su2018deeper}
J.-C. Su, M.~Gadelha, R.~Wang, and S.~Maji, ``A deeper look at 3d shape classifiers,'' in {\em Second Workshop on 3D Reconstruction Meets Semantics, ECCV}, 2018.

\bibitem{online_hamdi2022mvtn}
A.~Hamdi, F.~AlZahrani, S.~Giancola, and B.~Ghanem, ``Mvtn: Learning multi-view transformations for 3d understanding,'' 2022.

\end{thebibliography}

\end{document}


\title[MV-CLIP]{Supplementary Materials: \\MV-CLIP: Multi-View CLIP for Zero-shot 3D Shape Recognition}

\author{Anonymous Authors}

\maketitle

\section{Zero-Shot Performance for Each Class}

In this section, we present the zero-shot classification accuracy for each class in ModelNet40. As shown in Table \ref{tab:zero_shotper_class}, an increase in classification accuracy is observed in the majority of categories. It suggests that in most categories, such as laptop, bathtub, booksheld, etc., traditional multi-view setting contains a substantial amount of redundant information, which is detrimental for classification. The entropy minimization-based view selection is capable of extracting views with relatively clearer semantic content to represent 3D shapes. The hierarchical prompts also play a role in refining the classification results. 
For those categories such as airplane and person that almost all views contain distinguishable information, the zero-shot performance is already saturated.
Unfortunately, for some ambiguous categories such as flower pot and vase, hierarchical classification might adjust originally correct classifications to incorrect ones.

\input{table1/01}

\section{Sensitivity Analysis with Thresholds}

\begin{figure}[!h]
    \centering
    \includegraphics[width=0.75\linewidth]{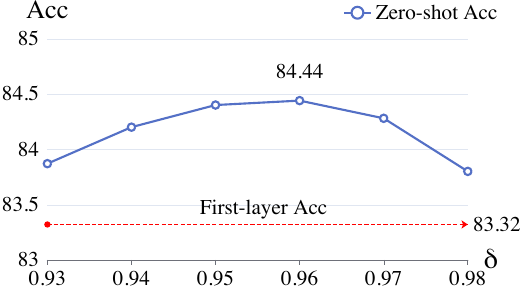}
    \caption{Sensitivity analysis with different confidence thresholds $\delta$ on ModelNet40. 
    }
    \label{fig:sensi}
\end{figure}

In this section, we conduct sensitivity analysis on different thresholds $\delta$ of hierarchical prompts module (introduced in Sec. 3.3.2 of the main manuscript). As shown in Figure~\ref{fig:sensi}, a lower threshold results in more high-confidence samples and less reliance on hierarchical matching, thus the accuracy tends to be closer to results without using hierarchical prompts. Conversely, a higher threshold increases the number of low-confidence samples.
It implies that for high-confident samples, the first-layer prompts work well and the improvement introduced by hierarchical prompts is limited. 
To balance accuracy with efficiency in hierarchical prompts matching, we set the threshold at 0.96 to refine the classification of unreliable samples. 

\section{Experiments with Offline Rendering}

\begin{figure}[t]
    \centering
    \includegraphics[width=0.9\linewidth]{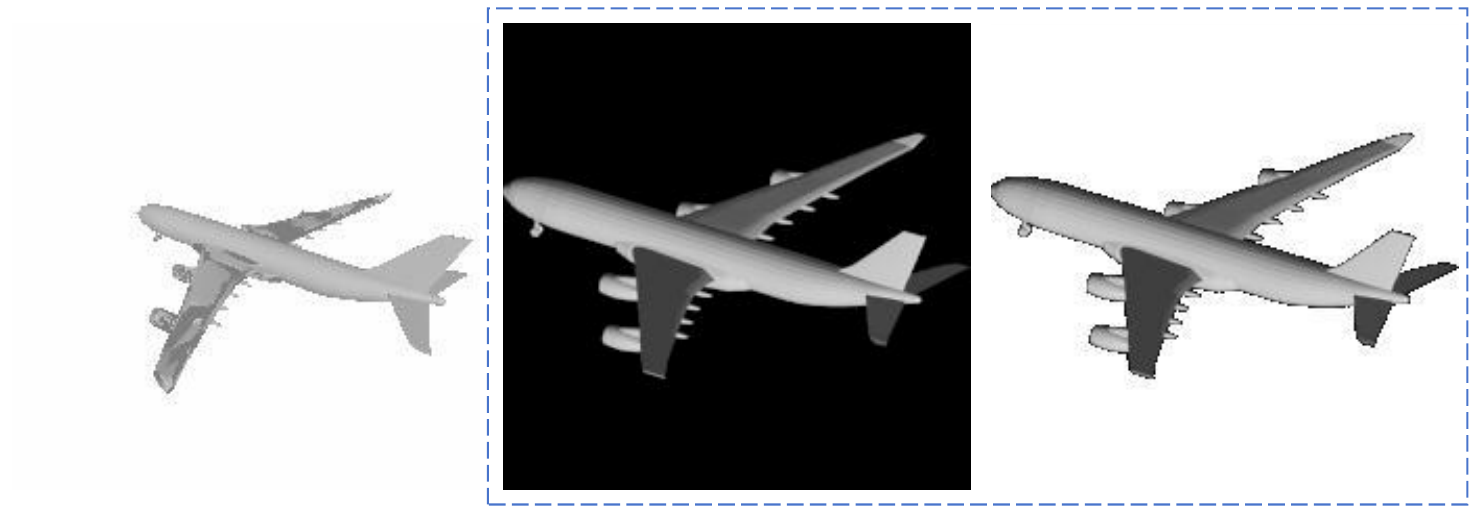}
    \caption{Comparison between online and offline rendered views. Left: an example of online rendering with sub-optimal rendering quality. Middle and right: an example of offline rendering with relatively better rendering quality.}
    \label{fig:offline}
\end{figure}

In this section, we present the experimental results obtained using offline views \cite{offline_su2018deeper}. As depicted in Figure~\ref{fig:offline}, views rendered offline exhibit higher quality with better shadow, but it suffers from slower rendering speed and storage requirement for 3D shapes. In contrast, online rendering views allow for timely operation and can cooperate with view selection to save storage. Therefore, we employ an online renderer \cite{online_hamdi2022mvtn} in our method and consider the results from the online renderer as the final outcome.

For consistency in experimental comparison, we alter the background color of the offline rendered views to white. Based on the experimental results shown in Table~\ref{tab:offline}, we observe that offline rendered views perform better than online ones. Furthermore, under offline view rendering, our approach achieves a consistent $4.83\%$ improvement, which substantiates the effectiveness of the modules we propose.

\begin{table}[t!]
    \centering
    \caption{ Zero-shot results on ModelNet40 (12 views) with different renderers. With a better renderer, the proposed modules still facilitate the performance.}
    \begin{tabular}{ccccc}
    \toprule
    \multirow{2}{*}{Renderer} & \multirow{2}{*}{Method}               & \multirow{2}{*}{$M$}     &\multirow{2}{*}{$M_{selec}$}    & \multirow{2}{*}{Acc} \\ 
                              &                                       &                             &                                   &                      \\  
    \midrule
    \multirow{5}{*}{Offline}  & \multirow{2}{*}{Ours}                 & \multirow{2}{*}{12}    & 6          & 85.05                \\
                              &                                       &                        & 4          & \textbf{85.17}       \\ \cline{2-5}
                              & \multirow{3}{*}{w/o designed modules} &                        & 12          & 80.34                \\
                              &                                       &                        & 6          & 76.78                \\
                              &                                       &                        & 4          & 64.16                \\
    \midrule
    \multirow{5}{*}{Online}   & \multirow{2}{*}{Ours}                 & \multirow{2}{*}{12}    & 6          & 82.53                \\
                              &                                       &                        & 4          & 82.49                \\ \cline{2-5}
                              & \multirow{3}{*}{w/o designed modules} &                        & 12          & 78.42                \\
                              &                                       &                        & 6          & 75.04                \\
                              &                                       &                        & 4          & 64.38                \\
    \bottomrule
    \end{tabular}
    \label{tab:offline}
\end{table}

\section{Viewpoint Configuration and Visual Results}
\begin{figure}[!ht]
    \centering
    \includegraphics[width=0.98\linewidth]{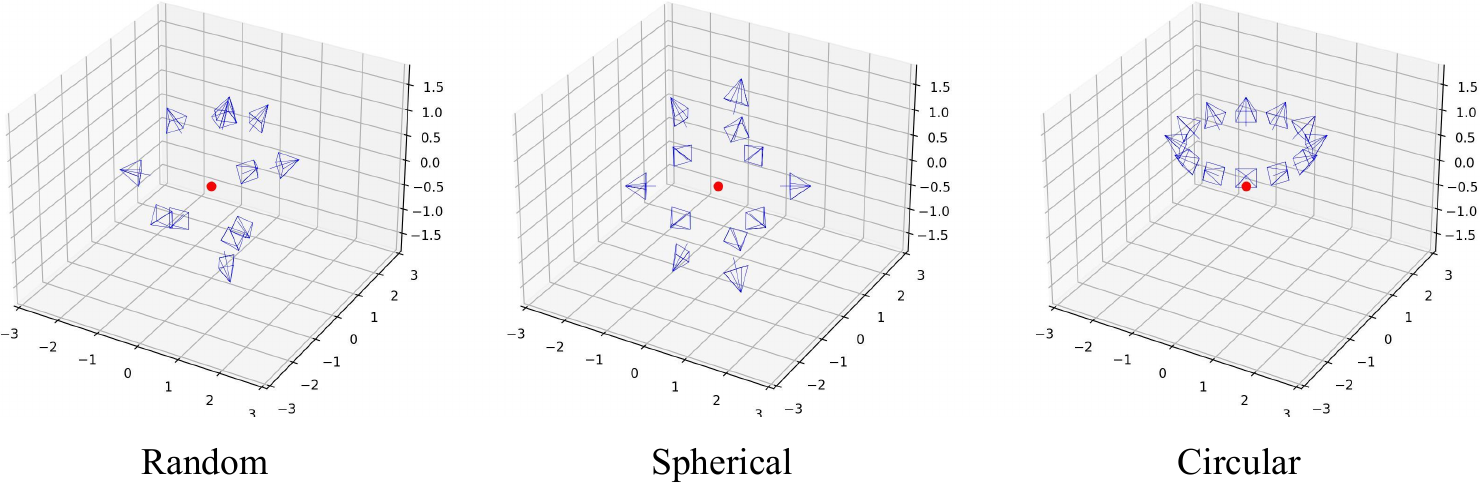}
    \caption{Multi-view camera configurations. The shape’s center is shown as a red dot, and the camera positions as blue cameras.}

    \label{fig:camera_positions}
\end{figure}

In this section, we present additional visualization results under the camera positions of random, spherical, and circular configurations, which is illustrated in Figure~\ref{fig:camera_positions}.

\begin{figure*}[t!]
    \centering
    \includegraphics[width=0.95\linewidth]{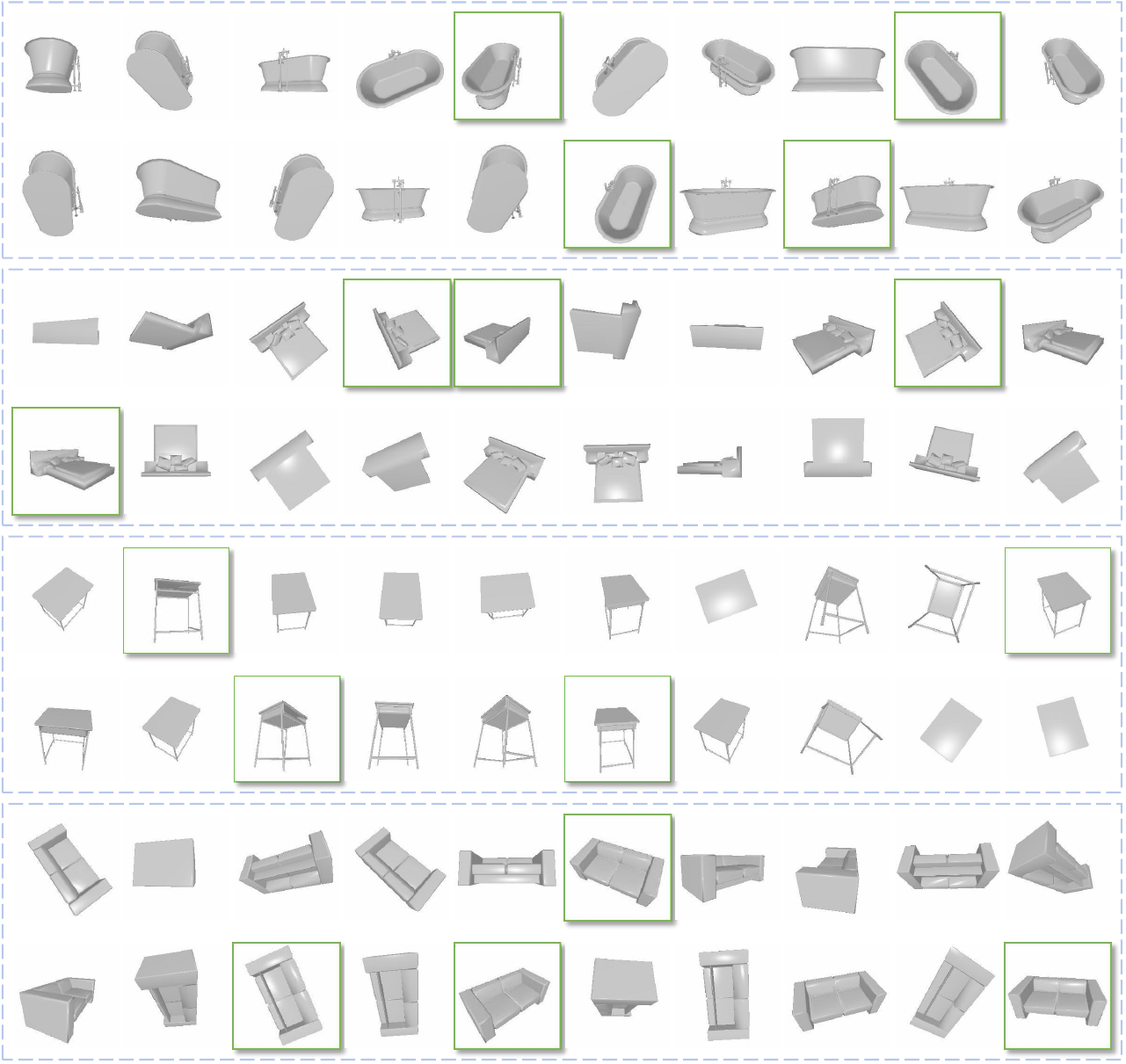}
    \caption{Visualization of multiple views from the models under the random camera positions. Note that the selected views are indicated by green boxes.}
    \label{fig:random}
\end{figure*}

\begin{figure*}[t!]
    \centering
    \includegraphics[width=0.95\linewidth]{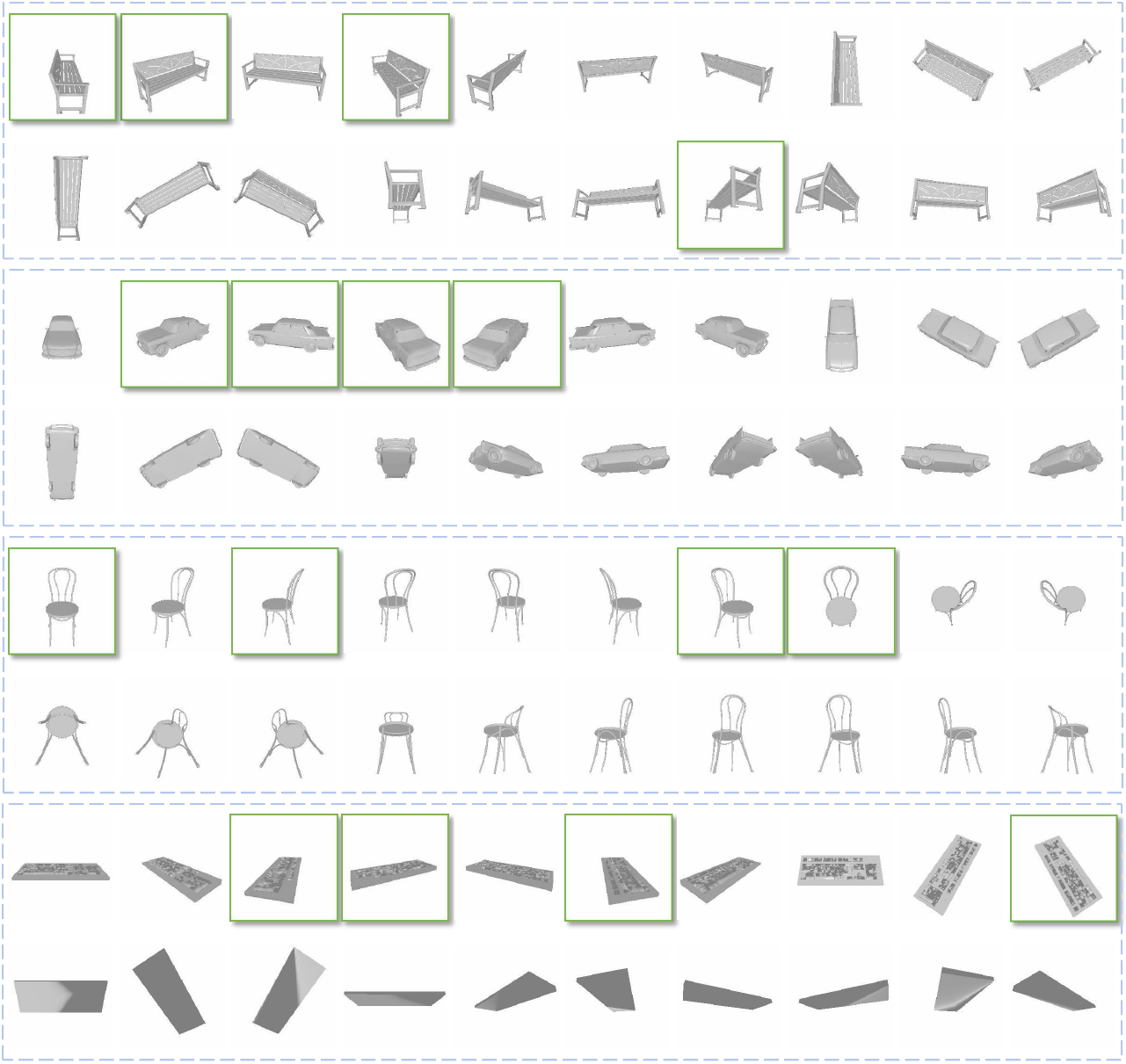}
    \caption{Visualization of multiple views from the models under the spherical camera positions. Under spherical camera configuration, views from the top and bottom of positions generally contain less semantic information, thus most of them are not selected. }
    \label{fig:spherical}
\end{figure*}

\begin{figure*}[t!]
    \centering
    \includegraphics[width=0.95\linewidth]{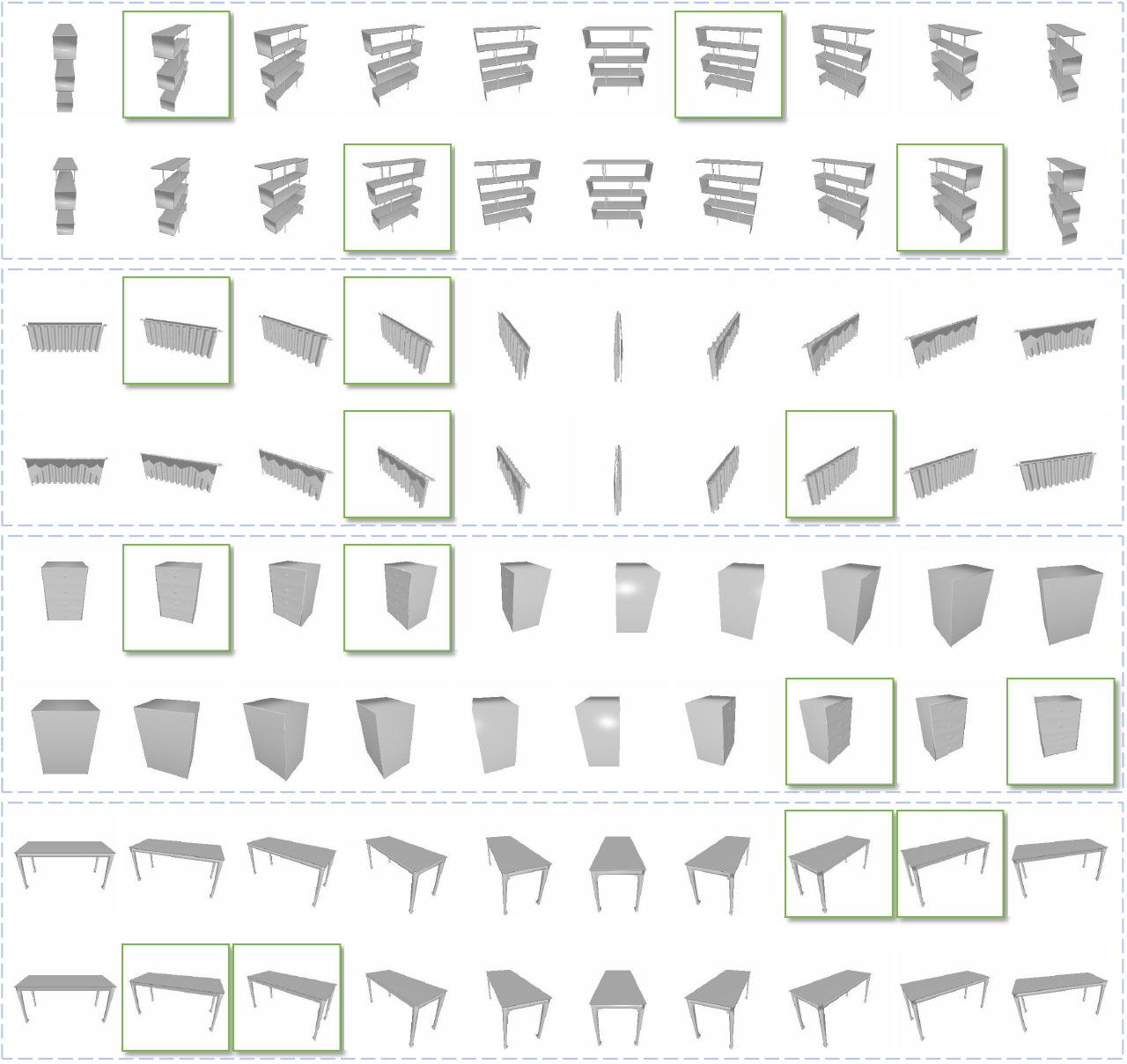}
    \caption{Visualization of multiple views from the models under the circular camera positions. Under circular camera configuration, the semantic information in the initial multiple views is generally clear, but the views that are selected have even more prominent semantics.}
    \label{fig:circular}
\end{figure*}

As shown in Figure~\ref{fig:random}, views obtained from random camera positions are quite disordered, with some views lacking distinct 3D shape characteristics. In such cases, view selection can effectively eliminate these disadvantageous views such as the bottom of a bathtub, the back and underside of a bed, the fan-shaped surface of a desk, and the back of a sofa, and select those with relatively clearer semantics. As illustrated in Figure~\ref{fig:spherical} and in Figure~\ref{fig:circular}, views obtained from spherical and circular camera positions are more comprehensive. However, there still exist some views that fail to represent 3D shapes effectively, where view selection remains beneficial.
Hence, if the semantic information in the initial multiple views is poorer, the impact of view selection is better.

\section{Limitation}

\begin{figure}[ht]
    \centering
    \includegraphics[width=1\linewidth]{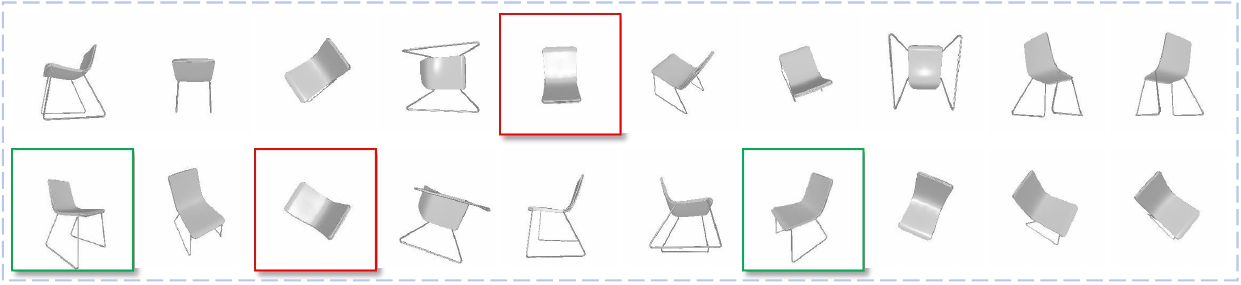}
    \caption{An example of incorrect view selection. The selected views exhibit semantic inconsistency and do not align with the ground truth.}
    \label{fig:limitation}
\end{figure}

The proposed MV-CLIP achieves state-of-the-art zero-shot 3D recognition performance using pre-trained vision-language models without any fine-tuning with 3D data, comparable to the most advanced technologies in the field. However, further exploration is needed in the following aspects. 
Firstly, some predefined settings such as the number of selected views and the number of candidate labels could be adaptive. 
Secondly, 
the selected views lack semantic consistency constraints. As illustrated in Figure~\ref{fig:limitation}, the views outlined in red are semantically identified as sink and laptop, which are inconsistent with the other views, i.e., chair.
Finally, the design of hierarchical prompts has limited effectiveness in improving classification for highly similar and difficult samples. For instance, as shown in Table \ref{tab:zero_shotper_class} for the flower pot and vase, the refined outcomes even incorrectly adjust originally correct results.

\bibliographystyle{ACM-Reference-Format}
\bibliography{sample-base}